\documentclass{article} % For LaTeX2e
\PassOptionsToPackage{table,svgnames,dvipsnames}{xcolor}
\usepackage{iclr2026_conference,times}

\usepackage{amsmath,amsfonts,bm}

\def\eqref#1{equation~\ref{#1}}
\def\1{\bm{1}}

\DeclareMathAlphabet{\mathsfit}{\encodingdefault}{\sfdefault}{m}{sl}
\SetMathAlphabet{\mathsfit}{bold}{\encodingdefault}{\sfdefault}{bx}{n}

\usepackage[utf8]{inputenc} % allow utf-8 input
\usepackage[T1]{fontenc}    % use 8-bit T1 fonts
\usepackage{hyperref}       % hyperlinks
\usepackage{url}            % simple URL typesetting
\usepackage{booktabs}       % professional-quality tables
\usepackage{amsfonts}       % blackboard math symbols
\usepackage{nicefrac}       % compact symbols for 1/2, etc.
\usepackage{microtype}      % microtypography

\usepackage{amsmath}
\usepackage{xfrac}
\usepackage{paralist}
\usepackage{multirow}
\usepackage{multicol}
\usepackage{enumitem}
\usepackage{subcaption}
\usepackage{makecell, cellspace}
\usepackage{relsize}
\usepackage{wrapfig}
\usepackage{graphicx}

\newcommand\ie{\textit{i.e.}}
\newcommand\eg{\textit{e.g.}}
\graphicspath{{./figures/}}

\title{Mixture-of-Experts Can Surpass Dense LLMs Under Strictly Equal Resource}

\newcommand{\boxwidthauthor}{5.1cm}

\author{
  \small
  \makebox[\boxwidthauthor][l]{Houyi Li$^{1,2}$} \\
  \makebox[\boxwidthauthor][l]{{\small\texttt{hyli22@m.fudan.edu.cn}}} \\
  \And
  \makebox[\boxwidthauthor][l]{Ka Man Lo$^{2}$} \\
  \makebox[\boxwidthauthor][l]{{\small\texttt{kamanphoebe@gmail.com}}} \\
  \AND
  \makebox[\boxwidthauthor][l]{Shijie Xuyang$^{1}$} \\
  \makebox[\boxwidthauthor][l]{{\small\texttt{ysjxu24@m.fudan.edu.cn}}} \\
  \And
  \makebox[\boxwidthauthor][l]{Ziqi Wang$^{3}$} \\
  \makebox[\boxwidthauthor][l]{{\small\texttt{wzq142857@mail.ustc.edu.cn}}} \\
  \AND
  \makebox[\boxwidthauthor][l]{Wenzhen Zheng$^{2}$} \\
  \makebox[\boxwidthauthor][l]{{\small\texttt{zhengwenzhen@amss.ac.cn}}} \\
  \And
  \makebox[\boxwidthauthor][l]{Haocheng Zhang$^{1}$} \\
  \makebox[\boxwidthauthor][l]{{\small\texttt{hczhang25@m.fudan.edu.cn}}} \\
  \AND
  \makebox[\boxwidthauthor][l]{Zhao Li$^{4}$} \\
  \makebox[\boxwidthauthor][l]{{\small\texttt{lzjoey@gmail.com}}} \\
  \And
  \makebox[\boxwidthauthor][l]{Shuigeng Zhou$^{1}$\thanks{Corresponding author. \quad $^\dagger$ \url{https://huggingface.co/kamanphoebe/moe_surpass_dense}.}}\\
  \makebox[\boxwidthauthor][l]{{\small\texttt{sgzhou@fudan.edu.cn}}} \\
  \AND
  \makebox[\boxwidthauthor][l]{Xiangyu Zhang$^{2}$} \\
  \makebox[\boxwidthauthor][l]{{\small\texttt{robert.zhang@stepfun.com}}} \\
  \And
  \makebox[\boxwidthauthor][l]{Daxin Jiang$^{2}$} \\
  \makebox[\boxwidthauthor][l]{{\small\texttt{djiang@stepfun.com}}} \\
  \AND
  \parbox{\linewidth}{\textnormal{
    $^1$Fudan University \quad $^2$StepFun \\[2pt]
    $^3$University of Science and Technology of China \quad $^4$Zhejiang University
  }}
}
\iclrfinalcopy % Uncomment for camera-ready version, but NOT for submission.
\begin{document}
% \raggedbottom

\maketitle

\begin{abstract}
Mixture-of-Experts (MoE) language models dramatically expand model capacity and achieve remarkable performance without increasing per-token compute. 
However, can MoEs \emph{surpass} dense architectures under strictly equal resource constraints --- that is, when the total parameter count, training compute, and data budget are identical?
This question remains under-explored despite its significant practical value and potential.
In this paper, we propose a novel perspective and methodological framework to study this question thoroughly.
First, we comprehensively investigate the architecture of MoEs and achieve an optimal model design that maximizes the performance.
Based on this, we subsequently find that an MoE model with activation rate in an optimal region is able to outperform its dense counterpart under the same total parameter, training compute and data resource.
More importantly, this optimal region remains consistent across different model sizes.
Although additional amount of data turns out to be a trade-off for enhanced performance, we show that this can be resolved via reusing data.
We validate our findings through extensive experiments, training nearly 200 language models at 2B scale and over 50 at 7B scale, cumulatively processing 50 trillion tokens. 
All model checkpoints are publicly available$^\dagger$.
\end{abstract}

% Nevertheless, current MoE models tend to underperform their dense equivalents, possibly because the conditional computation within MoE makes their training challenging.
% It remains unsure whether the MoE architecture can be consistently superior to conventional dense models.  
\section{Introduction}
\label{sec:intro}
In recent years, Large Language Models (LLMs) built on the Transformer architecture~\citep{vaswani2017attention} have achieved strong results on a range of NLP tasks~\citep{radford2018improving, achiam2023gpt, touvron2023llama1, touvron2023llama2, bai2023qwen}. 
Meanwhile, there has been growing interest in using Mixture-of-Experts (MoE) layers~\citep{shazeer2017outrageously} to expand model capacity while keeping the training cost reasonable~\citep{switchtransformers, stmoe, deepspeedmoe}.
Recent open-source initiatives have explored MoE-based LLMs~\citep{deepseekmoe, mixtralmoe, wei2024skywork, xue2024openmoe, deepseekv3}, yet many widely adopted open-source models such as LLaMA~\citep{touvron2023llama1, touvron2023llama2}, DeepSeek's first-generation models~\citep{deepseekv1}, and Qwen~\citep{yang2024qwen2technicalreport, Qwen2p5}, continue to utilize dense architectures, posing an open question of \emph{whether MoE LLMs can outperform their dense counterparts}.

Current comparisons of MoE and dense LLMs often simplify the analysis to either a \emph{data-centric} or a \emph{compute-centric} perspective. 
The data-centric view, which keeps total training tokens constant for both MoE and dense models, praises MoE for its reduced activated parameter count per token (i.e., per-token compute cost) and potential for aggressive parameter scaling.
For example, DeepSeekMoE~\citep{deepseekmoe} reports a 16B-parameter MoE model (with 2.5B activated parameters) that achieves performance on par with a 7B dense model under the same data budget, thus suggesting a 2.5$\times$ ``parameter-efficiency'' advantage. 
The compute-centric view, which fixes total training compute, examines the impact of MoE sparsity (\ie, the ratio of activated to total parameters) on performance.
Under certain sparse configurations, total parameters can swell to nearly 100$\times$ those of a dense baseline, but at the cost of requiring more training data and managing high memory overhead.

% \paragraph{Limitations of Existing Approaches} 
However, neither perspectives fully addresses the complex interplay of critical resource constraints in large-scale model development: the finite nature of \textbf{training data volume} ($D$), \textbf{training compute} ($C$), and \textbf{model size} ($N$), which affects both memory and inference throughput. 
In particular, MoE models typically encounter bandwidth bottlenecks during inference, as all experts reside in GPU high-bandwidth memory and must be moved into shared memory, making parameter count a key runtime cost factor beyond FLOPs.
These interdependencies complicate the conclusion of the absolute superiority of MoE or dense architectures. 
Intuitively, a dense model with equivalent total parameters should have an advantage by fully utilizing its capacity. 
This often leads studies to favor MoE for scaling model size rather than direct comparisons at the same parameter count,  thus overlooking the real-world resource constraints in large-scale training and deployment.

In this work, we introduce a novel perspective aimed at providing a more convincing resolution to this debate by posing the following question:
\begin{quote} 
    \emph{Can Mixture-of-Experts surpass dense LLMs under equal total parameter, compute, and data constraints?}
\end{quote}

To reach a definitive conclusion on this issue, we draw insights from a unified parameterization framework for model architecture (\S~\ref{sec:method}) and propose a three-step experimental methodology.
\textit{First}, we search for an optimized architecture design to ensure each model candidate achieves its (near-)~optimal performance (\S~\ref{sec:arc}).
\textit{Second}, we explore the optimal activation rate based on this optimized model architecture, with the total parameters $N$ and compute budget $C$ fixed (\S~\ref{sec:activation_rate}).
\textit{Third}, we present a data reuse strategy to address the additional data demand of MoE models, thereby equating data resource $D$ (\S~\ref{sec:data_reuse}).
We also analyze the efficacy of this framework on downstream tasks (\S~\ref{sec:downstream}).

Our findings, derived from extensive experiments and systematic evaluation under the proposed strict $N$/$C$/$D$ parity with dense models, provide strong evidence that \textit{MoE architectures with optimized backbones and activation rates can indeed achieve superior performance over dense models on both upstream and downstream tasks}.
This implies that any observed performance gains can be attributed solely to architectural advantages, rather than disparities in parameter count or compute budget.
Moreover, this challenges conventional wisdom and paves the way for resource-efficient yet powerful architectures in the next generation of large-scale NLP systems.

Our main contributions are:
%\begin{itemize} 
%    \item %\textbf{Outperforming dense models at equal size:} 
    1) We demonstrate, for the first time, that under fixed total parameters ($N$) and a fixed compute budget ($C$), an MoE LLM can surpass its dense counterpart with careful architecture design (see Fig.~\ref{fig:2B_optimal_ra_keepC}, \ref{fig:7B_optimal_ra_keepC}). 
    %\item %\textbf{Optimal activation rate (AR) region:} 
    2) Our experiments reveal the existence of a stable ``optimal AR'' region that consistently maximizes performance across varying $N$ (see Fig.~\ref{fig:2B_optimal_ra}, \ref{fig:7B_optimal_ra}).
    %\item %\textbf{Data efficiency via reuse:} 
    3) We introduce a practical data reuse strategy that offsets MoE's additional data needs, enabling robust gains over dense models without substantially increasing unique training data $D$ (see Fig.~\ref{fig:7B_optimal_ra}, \ref{fig:downstream}).
\section{Related Work}

\subsection{MoE Language Models}

% In the context of deep learning, sparsely-gated MoE layer was first introduced by \citet{shazeer2017outrageously} on recurrent language models, enabling dramatic scaling in model capacity with modest losses in computational efficiency.
Building upon MoE~\citep{shazeer2017outrageously}, GShard~\citep{lepikhin2020gshard} facilitates model parallelism across devices for massive MoE models.
With the advent of Transformers~\citep{vaswani2017attention}, the integration of MoE into the Transformer framework has become a popular model architecture and achieved state-of-the-art performance.
As an early attempt, Switch Transformer~\citep{switchtransformers} proposes top-1 gating to simplify MoE architecture and alleviate communication overhead. 
More recent Transformer-based MoE LLMs include~\citep{xue2024openmoe, mixtralmoe, wu2024yuan, wei2024skywork, deepseekv3}. 
% To integrate the Mixture-of-Experts (MoE) architecture into Transformer models, a common implementation is to substitute the feed-forward network (FFN) with an MoE block that consists of a gate and a number of experts.
The MoE architecture is briefly reviewed in Appendix~\ref{append:background}.
% Intuitively, only experts that are assigned the highest $k$ scores by the gate will be chosen to process the current input. 

% Despite their successes, MoE models face challenges such as load balancing among experts and increased complexity in training dynamics. Strategies like auxiliary loss functions for balancing expert utilization and improved routing mechanisms have been proposed to address these issues.
% \citet{cai2024survey} provides a comprehensive review of both the algorithmic and system design aspects of MoEs.

% TODO 
\subsection{Analyses of MoE Sparsity}

Several studies investigated the impact of varying the number of MoE experts and adjusting granularity, both of which are factors related to sparsity.
Through ablation studies, DeepSeekMoE~\citep{deepseekmoe} observes finer granularity results in improvement on overall model performance, and acquires a ratio between shared and routed experts that yields slightly better Pile loss. 
\citet{stmoe} summarized the results of several MoE works and indicated that the gain of increasing sparsity quickly diminishes when the number of experts is greater than 256 --- a very sparse model.

% \begin{wraptable}{r}{5.55cm}
% % \begin{table}[thb]
%     % \centering
%     \caption{Notation.}
%     \label{tab:part_notation}
%     \footnotesize
%     \setlength\tabcolsep{3pt}
%     \begin{tabular}{cl}
%         \toprule
%         Symbol & Definition \\
%         \midrule
%         % $\mathcal{L}$ & Cross entropy loss. \\
%         % $\hat{N}$ & Total number of parameters. \\
%         $L_\text{e}$ & Number of MoE layers. \\
%         $L_\text{d}$ & Number of dense layers. \\
%         % $L$ & Number of total layers, \ie, $L_\text{e} + L_\text{d}$. \\
%         $L$ & Number of total layers \\
%         $S$ & Sequence length. \\
%         $D_\text{m}$ & Model hidden dimension. \\
%         $D_\text{ffn}$ & FFN hidden dimension. \\
%         $D_\text{e}$ & Expert hidden dimension. \\
%         $D_\text{se}$ & Shared expert hidden dimension. \\
%         % \ie, $K \cdot D_\text{e}$. \\
%         $E$ & Number of experts. \\
%         $K$ & Number of chosen experts. \\
%         \bottomrule
%     \end{tabular}
% \end{wraptable}

\begin{table}[thb]
    \centering
    \caption{Notation.}
    \label{tab:notation}
    \footnotesize
    \setlength{\tabcolsep}{6pt}
    \newlength{\notationdefwidth}
    \begin{minipage}[t]{0.49\columnwidth}
        \setlength{\notationdefwidth}{0.78\linewidth}
        \centering
        \begin{tabular}{@{}c l@{}}
            \hline
            Symbol & Definition \\
            \hline
            $D$ & \parbox[t]{\notationdefwidth}{Dataset size in tokens.} \\
            $M$ & \parbox[t]{\notationdefwidth}{Compute (w/o embedding) per token in FLOPs.} \\
            $C$ & \parbox[t]{\notationdefwidth}{Total training compute in FLOPs, \ie, $M \cdot D$.} \\
            $N$ & \parbox[t]{\notationdefwidth}{Number of non-vocabulary parameters.} \\
            $N_\text{a}$ & \parbox[t]{\notationdefwidth}{Number of activated parameters.} \\
            $r_\text{a}$ & \parbox[t]{\notationdefwidth}{Activation rate, \ie, $\sfrac{N_\text{a}}{N}$.} \\
            $L_\text{e}$ & \parbox[t]{\notationdefwidth}{Number of MoE layers.} \\
            $L_\text{d}$ & \parbox[t]{\notationdefwidth}{Number of dense layers.} \\
            $L$ & \parbox[t]{\notationdefwidth}{Number of total layers, \ie, $L_\text{e} + L_\text{d}$.} \\
            $\alpha$ & \parbox[t]{\notationdefwidth}{FFN expansion ratio, \ie, $\sfrac{D_\text{ffn}}{D_\text{m}}$.} \\
            $\zeta$ & \parbox[t]{\notationdefwidth}{Model aspect ratio, \ie, $\sfrac{D_\text{m}}{L}$.} \\
            $\gamma$ & \parbox[t]{\notationdefwidth}{Sequence-to-width ratio, \ie, $\sfrac{S}{D_\text{m}}$.} \\
            \hline
        \end{tabular}
    \end{minipage}
    \hfill
    \begin{minipage}[t]{0.49\columnwidth}
        \setlength{\notationdefwidth}{0.78\linewidth}
        \centering
        \begin{tabular}{@{}c l@{}}
            \hline
            Symbol & Definition \\
            \hline
            $S$ & \parbox[t]{\notationdefwidth}{Sequence length.} \\
            $H$ & \parbox[t]{\notationdefwidth}{Number of attention heads.} \\
            $D_\text{m}$ & \parbox[t]{\notationdefwidth}{Model hidden dimension.} \\
            $D_\text{ffn}$ & \parbox[t]{\notationdefwidth}{FFN hidden dimension.} \\
            $D_\text{h}$ & \parbox[t]{\notationdefwidth}{Dimension of attention head.} \\
            $D_\text{e}$ & \parbox[t]{\notationdefwidth}{Expert hidden dimension.} \\
            $D_\text{se}$ & \parbox[t]{\notationdefwidth}{Shared expert hidden dimension.} \\
            $E$ & \parbox[t]{\notationdefwidth}{Number of experts.} \\
            $K$ & \parbox[t]{\notationdefwidth}{Number of chosen experts.} \\
            $\beta$ & \parbox[t]{\notationdefwidth}{Activated FFN-to-model ratio in MoE layers, \ie, $\sfrac{(D_\text{se}+KD_\text{e})}{D_\text{m}}$.} \\
            $\mu$ & \parbox[t]{\notationdefwidth}{Total FFN-to-model ratio in MoE layers, \ie, $\sfrac{(D_\text{se}+ED_\text{e})}{D_\text{m}}$.} \\
            \hline
        \end{tabular}
    \end{minipage}
\end{table}

From a methodological perspective, our comparisons are conducted in a sufficiently trained regime (with $\sfrac{D}{N}\ge 20$ for key models) 
and under a fixed-$N$ setting motivated by deployment memory constraints, 
which differs from scaling-law sweeps that often rely on undertrained large models at limited compute. 
We provide a more detailed discussion in Appendix~\ref{append:related_work}.

% One line of scaling law researches focuses on MoE models.
% \citet{krajewski2024scaling} established a scaling law for fine-grained MoE, which reveals that higher granularity values are able to enhance MoE model performance and efficiency.
% While \citet{clark2022unified} drew a similar conclusion regarding granularity, their proposed scaling laws are applicable to both dense models and different routing models.
Concurrently with our work, \citet{ludziejewski2025jointmoescalinglaws} found that a sufficiently large MoEs trained with more tokens outperforms a dense model with the same total parameters. 
We further show MoE superiority even at smaller sizes and address the additional data demand via reuse.
\citet{abnar2025parameters} studied the scaling law for optimal MoE sparsity. 
However, their models (up to $N=30\text{B}$) were trained with $C=1\mathrm{e}20$, a much smaller budget compared to the  approximately $9\times$ and $30\times$ compute we used for our 2B and 7B models to ensure an adequate $\sfrac{D}{N}$.
This likely resulted in undertrained models, potentially affecting their conclusions.
Moreover, our study first optimizes the MoE architecture to isolate the effect of different activation rates on performance.
% we examine ``sparsity'' in a more general sense, considering various MoE architectural designs, and derive different conclusions from most previous works. 
% For instance, we identify an optimal activation rate region that is independent of model sizes. 
Detailed differences between our work and this previous study are discussed in Appendix~\ref{append:related_work}.

\section{Experimental Methodology}
\label{sec:method}

We begin by introducing a unified parameterization framework for model architecture, establishing a solid foundation.
Then, we derive key insights from this parameterization, which informs our comprehensive three-step experimental methodology. 
Finally, we detail the experimental setup used consistently across all subsequent experiments. Our notation is summarized in %Table~\ref{tab:part_notation} and 
Table~\ref{tab:notation}.

\subsection{Architecture Parameterization}

% 为了全面的、一般性的对比Dense LLM架构和MoE LLM架构，我们首先需要把两种架构参数化。和之前工作不同的是，我们考虑了工业使用场景中的一般MoE架构，即MoE允许部分层（不是全部层）是Dense结构。
% 为了表述清楚，我们将模型结构的各个指标的数学Notation放在了表格\ref{tab:notation}中。
% 对于Dense模型其非词表参数量和per token计算量简单近似为：
To enable a comprehensive and general comparison of dense and MoE-based LLM architectures under realistic deployment constraints, we introduce a unified parameterization framework that explicitly accounts for both model parameters and per-token compute cost.

\paragraph{Dense model parameterization.}
For a dense model, we approximate the number of non-embedding parameters $N$ and the per-token forward-pass computation cost $M$ as follows:
\begin{align}
N \approx{}& (4+3\alpha) D_\text{m}^2 L = (4+3\alpha) \zeta^2 L^3, \label{eq:dense_n}\\
M \approx{}& 2N+4D_\text{m}SL = 2N + 4\zeta^2\gamma L^3 \label{eq:dense_m}\\
  ={}&       2N(1+\sfrac{2\gamma}{(4+3\alpha)}) \label{eq:dense_mm},
\end{align}
where $\alpha ={} \sfrac{D_\text{ffn}}{D_\text{m}}, \gamma = \sfrac{S}{D_\text{m}}, \text{ and } \zeta = \sfrac{D_\text{m}}{L}$.
Here, we omit the LayerNorm parameter count as it is negligible. 
Inference cost or training cost can be approximated by $C \approx M \times D$ or $C \approx 3M \times D$, respectively (based on the standard empirical observation that training typically takes about three times the forward pass for backward computation). 

\paragraph{MoE model parameterization.}
In many real-world settings, only a subset of Transformer layers are replaced with MoE layers.
The approximations for total non-vocabulary parameters $N$, activated parameters $N_\text{a}$, and per-token computation cost $M$ can be expressed as
% 而对于MoE模型来说其非词表参数量和per token计算量简单近似为：
\begin{align}
    \begin{split}
        \label{eq:moe_n}
        N \approx{}& (4+3\mu)D_\text{m}^2L_\text{e} + (4+3\alpha)D_\text{m}^2L_\text{d},
    \end{split}\\
    \begin{split}
        \label{eq:moe_na}
        N_\text{a} \approx{}& (4+3\beta)D_\text{m}^2L_\text{e} + (4+3\alpha)D_\text{m}^2L_\text{d},
    \end{split}\\
    \begin{split}
        \label{eq:moe_m}
        M \approx{}& N_\text{a} + 4D_\text{m}SL = 2r_\text{a} N + 4\zeta^2\gamma L^3,
    \end{split}
\end{align}
where $\mu ={} \sfrac{(D_\text{se} + ED_\text{e})}{D_\text{m}} \text{ and } \beta ={} \sfrac{(D_\text{se} + KD_\text{e})}{D_\text{m}}$. 
We again omit the parameters and FLOPs of the gating network (router) as they are comparatively small, and the \emph{activation rate} (AR) is denoted as $r_\text{a}$.
% 这里我们也同时忽视了router的参数量和计算量，由于他们的占比很小。在上述公式中，我们分别对dense、MoE模型的参数量、计算量进行了解构和参数化，其中：
For the simple and common case where \emph{all} $L$ layers are MoE layers (\ie, $L_{\text{d}}=0$), we have
% 不妨考虑一种简单的MoE结构，$L_d=0$，同时这种情况也是极为常见的方案，即LLM中所有的层都是MOE层。此时MOE模型的per token的计算复杂度可以进一步写成：
\begin{align}
    r_\text{a} ={}& N_\text{a} / N = \sfrac{(4+3\beta)}{(4+3\mu)}, \\
    M \approx{}& 2r_\text{a} N + 4\zeta^2\gamma L^3  \\
      % = &      2r_\text{a}N + 4\zeta^2\gamma\frac{N}{(4+3\mu)\zeta^2} \\
      % % = &      2r_\text{a} N + \frac{4\gamma}{4+3\mu} N \\
      % = &      2r_\text{a} N + \frac{4\gamma r_\text{a}}{4+3\beta} N \\
      ={}&      2r_\text{a} N (1+\sfrac{2\gamma}{(4+3\beta)}). 
    \label{eq:moe_mm}
\end{align}

% 基于上述的对Dense Based LLM和MoE Based LLM的解构和参数化，我们进行以下分析，并确定本文的方法论。
\subsection{Key Observations and Methodological Considerations}
\label{subsec:method_consideration}

Based on the above parameterization, we highlight the following insights that guide our experiments:

% 第一，相较于Dense模型来说的MoE LLM的结构自由度更高。例如对于给定N/S的Dense LLM来说，宽高比$\zeta$和FNN倍数$\alpha$就可以唯一确定模型的shape和计算量。有一些之前的工作明确了Dense LLM最优Performance的$\zeta, \alpha$取值。
% 而对于MoE LLM来说，不仅先需要考虑到$L_{d},L_{e}$的分配。就算在$L_{d}=0$的情况下，$\zeta, \mu, r_\text{a}$仅可以确定大体的形状，具体的形状还要受限于$K,E,D_e,D_s$对一层FFN参数的分配方式。
% 由于MoE的结构自由度过高，如果通过网格搜索所有自由度的组合，过于花费算力。本文采取的方法，是优先梳理出MoE结构自由度的最小单元，每一步中的每一组实验都是基于之前所有实验的最优最优配置再进行下一步的消融对比，从而贪心的求解出最优的结构。

\paragraph{High structural degrees of freedom in MoE.}
Compared to a dense model (whose shape is almost uniquely determined by $L$, $D_\text{m}$, and $\alpha$), an MoE model has many more design choices: the number of MoE layers ($L_\text{e}$), expert-related dimensions (\eg, $K$, $E$, $D_{\text{e}}$, $D_{\text{se}}$), and so forth. 
Even with $L_\text{d}=0$, the final shape depends on $\mu$ and $\beta$ in addition to $\zeta$ and $r_\text{a}$. 
Exhaustively searching all combinations is prohibitively expensive. 
Therefore, a greedy strategy should be adopted.
% 第二，影响相同总参数量下的MoE LLM的Dense LLM的per token计算量之比的最主要因素为Activation rate$r_\text{a}$，次要因素为$\beta$，如公式\ref{eq:dense_mm}和\ref{eq:moe_mm}. 
% 不妨将同总参数量下的MoE LLM的计算复杂度相较于Dense LLM的节省比例计作$R_c$可以写作入下：
\paragraph{Activation rate $r_\text{a}$ is the primary factor.}
At the same total parameter count $N$, the ratio of per-token FLOPs between an MoE model and a dense model is primarily determined by the activation rate $r_\text{a}$. 
More specifically, if $M_\text{d}$ is the per-token cost of the dense baseline (with $\zeta,\alpha$ fixed), then the pure MoE model's compute, normalized by $M_\text{d}$, roughly behaves as (the union of Equ.~\ref{eq:dense_mm} and~\ref{eq:moe_mm}):
\begin{align}
    R_\text{c} &=r_\text{a}(\frac{4+3\alpha+2{\gamma}_\text{d}}{4+3\beta+2{\gamma}_\text{m}}) 
    \label{eq:ratio}
\end{align}
% 其中${\gamma}_d,{\gamma}_m$分别是Dense LLM和MoE LLM的长宽比。sequence length S是一个应用需求参数，往往是给定的，$\gamma$此时受$D_m$的影响和$\zeta$存在内在联系性。
% $\zeta,\alpha,\beta$都是结构相关的参数，在确定在Dense和MoE最优的$\zeta,\alpha,\beta$取值的情况下，$R_c$是随着$r_\text{a}$单调递增的。
where $\gamma_{\text{d}}$ and $\gamma_{\text{m}}$ denote $\sfrac{S}{D_\text{m}}$ for dense and MoE layers, respectively. 
As $\gamma$ strongly correlates with $\zeta$, once the shape hyperparameters $(\zeta,\alpha,\beta)$ are chosen, $R_\text{c}$ grows monotonically with $r_\text{a}$.

% 第三，当给定N能确定Dense/MoE LLM近似最优的结构的情况下，MoE总的训练计算量可以写作$C=3 R_c M_d D$，其中$M_d$为总参一致的Dense模型的Per Token计算量。不难发现如果同时保持MoE LLM计算量与Dense LLM训练总计算计算量一致，MoE模型需要$R_c$倍的数据。
\paragraph{The trade-off among $N$, $C$, and $D$.}
Once $N$ (the total parameters) is chosen and we fix near-optimal shapes for dense and MoE models, the total training compute for the MoE model can be approximated by $C = 3 \, R_\text{c} \, M_\text{d} \, D$, where $M_\text{d}$ is the per-token cost of the dense baseline with the same total parameter count, and $R_\text{c}$ is the fraction by which the MoE model reduces compute per token (relative to the dense baseline). 
If we want to keep the same total compute $C$ for both MoE and dense models, the MoE model will need $R_\text{c}$ times more training tokens.
% (i.e., $D$ must increase by a factor of $R_c$).

% 综合上述分析，本文首次提出了一种全新的方法论：即在保持C/D/N一致这一新的研究视角下的，可以实现Dense LLM和MoE LLM的全面公平对比。
% 1，用贪心的方法确定MoE的最优结构。1a, 最先确定MoE层($L_e$)的比例，这样可以将后续的分析聚焦于MoE本身的内部。1b, 再确定MoE层FFN参数的分配方式，他们是独立于$\zeta, \beta$的模型结构自由度。1c, 最终确定MoE模型下的最优$\zeta,\beta$。
% 2. 在都采用了近似最优结构的基础上，研究相同参数、计算量的下的Dense LLM和不同$r_\text{a}$MoE LLM的对比关系。此时MoE LLM需要更多的数据量。
% 3. 研究MoE LLM数据复用策略，使得他和Dense LLM的unique Data的消耗量一直，从而达到和Dense LLM全面公平的对比。

\subsection{Three-Step Experimental Methodology}
Motivated by the aforementioned observations, we propose a three-step experimental methodology which is both comprehensive and fair, so as to achieve the \textbf{new perspective} motivated in \S~\ref{sec:intro} that enables a more conclusive comparison of MoE and dense LLMs under equal resource constraints.
The methodology is outlined as follows:
% Specifically,  and \emph{enforce} that total parameters ($N$), total compute ($C$), and effective data ($D$) remain the same between the two models. This leads to three key steps:
%
% \begin{enumerate}[leftmargin=*]
%\begin{enumerate}
%    \item 
    1) \textbf{Greedy architecture determination.} 
    First, determine the macro-level layer composition (\ie, the MoE-to-dense layer ratio $L_\text{e}$ \textit{vs.}\ $L_\text{d}$ and related choices such as shared experts).
    Second, determine the micro-level MoE design within each MoE layer (\eg, top-$K$ routing and parameter allocation among routed/shared experts), which is largely orthogonal to the global model shape.
    Finally, select the near-optimal shape hyperparameters (e.g., $\zeta, \alpha$ for dense and $\zeta, \beta$ for MoE) for fair comparisons at a fixed $N$.
 %   \item 
    2) \textbf{Activation rate analysis under fixed $N$ and $C$.} 
    With the optimal MoE architecture chosen in the previous step, and the optimal dense LLM shape proposed by~\citet{oai_scaling}, we compare MoE models versus a dense baseline of the same size, ensuring the total training compute $C$ is matched. 
    Since $C$ must be the same, the MoE model typically receives up to $R_\text{c}$ times more tokens (initially considering repeated or augmented data).
%    \item 
    3) \textbf{Data reuse strategy.} 
    To ensure a \emph{truly} fair comparison at the same \emph{unique} data budget $D$, we develop a data reuse strategy that offsets MoE's additional data requirement. 
    This enables evaluation under strictly equal $N$, $C$, and $D$.
%\end{enumerate}

\subsection{Common Experimental Setup}
\label{sec:exp_setup}

% To ensure the fairness of comparisons between our trained models, we have worked on different aspects including model architecture and training hyperparameters, which will be introduced in this section and the following section.
% \textit{Under these common configurations, all models are expected to be trained properly and to achieve their almost optimal performance.}

% \noindent\textbf{Optimized Architecture.}
% Our experiments of investigating the optimal activation rates (ARs) depend on optimized backbones.
% In \S~\ref{sec:arc}, we conduct a series of experiments regarding several MoE model components (\eg, layer arrangement and shared expert) to examine the optimal architecture.
% Based on the findings, we design model backbones for the subsequent trainings in \S~\ref{sec:activation_rate} and discover optimal ARs under the condition of these optimized backbones.

\noindent\textbf{Optimal hyperparameters.}
MoE training is sensitive to the learning rate ($\eta$) and batch size ($B$)~\citep{he2024upcycling}. 
Even minor architectural changes, such as variations in $E$, can lead to different optimal hyperparameters. 
To address this, we train all our models using the optimal $\eta$ and $B$ based on the hyperparameter scaling laws proposed in \citep{li2025predictable}. 
Specifically, \citet{li2025predictable} found that the optimal $\eta$ and $B$ follow power laws and depend only on $N$ and $D$.
Since these scaling laws are applicable to both dense and MoE models and are robust across various pretraining data distributions, we apply them to determining $\eta$ and $B$ for each of our experiments.

\noindent\textbf{Dense baseline tuning.}
To ensure a strictly fair comparison, we also tuned the dense baselines by searching for near-optimal structural ratios (e.g., aspect ratio $\zeta$ and FFN expansion $\alpha$, equivalently $L, D_\text{m}, D_\text{ffn}$ with given N), guided by the scaling-law recommendations in \citet{li2025predictable}; the resulting dense configurations used throughout the paper are summarized in Table~\ref{tab:dense_baseline}.

\noindent\textbf{Others.}
We use internal, high-quality training and validation datasets composed primarily of diverse web text and specific domains such as mathematics and code. 
The training and validation sets have different distributions, requiring the evaluated models to demonstrate strong generalization capabilities. 
Our models incorporate RMSNorm~\citep{zhang2019root} for pre-normalization, ALiBi~\citep{press2021train} positional encoding for multi-head attention, and the SwiGLU~\citep{shazeer2020glu} activation function for both feed-forward networks (FFNs) and MoE experts. 
The training procedures used consistently across all experiments are outlined in Table~\ref{tab:training_recipe} in Appendix~\ref{appedix-results}. 
We employ cross-entropy loss ($\mathcal{L}$) as the training metric and bits-per-character (BPC) as the validation metric.

\begin{figure*}[t!]
    \centering
    \begin{subfigure}{0.44\linewidth}
        \includegraphics[width=\linewidth]{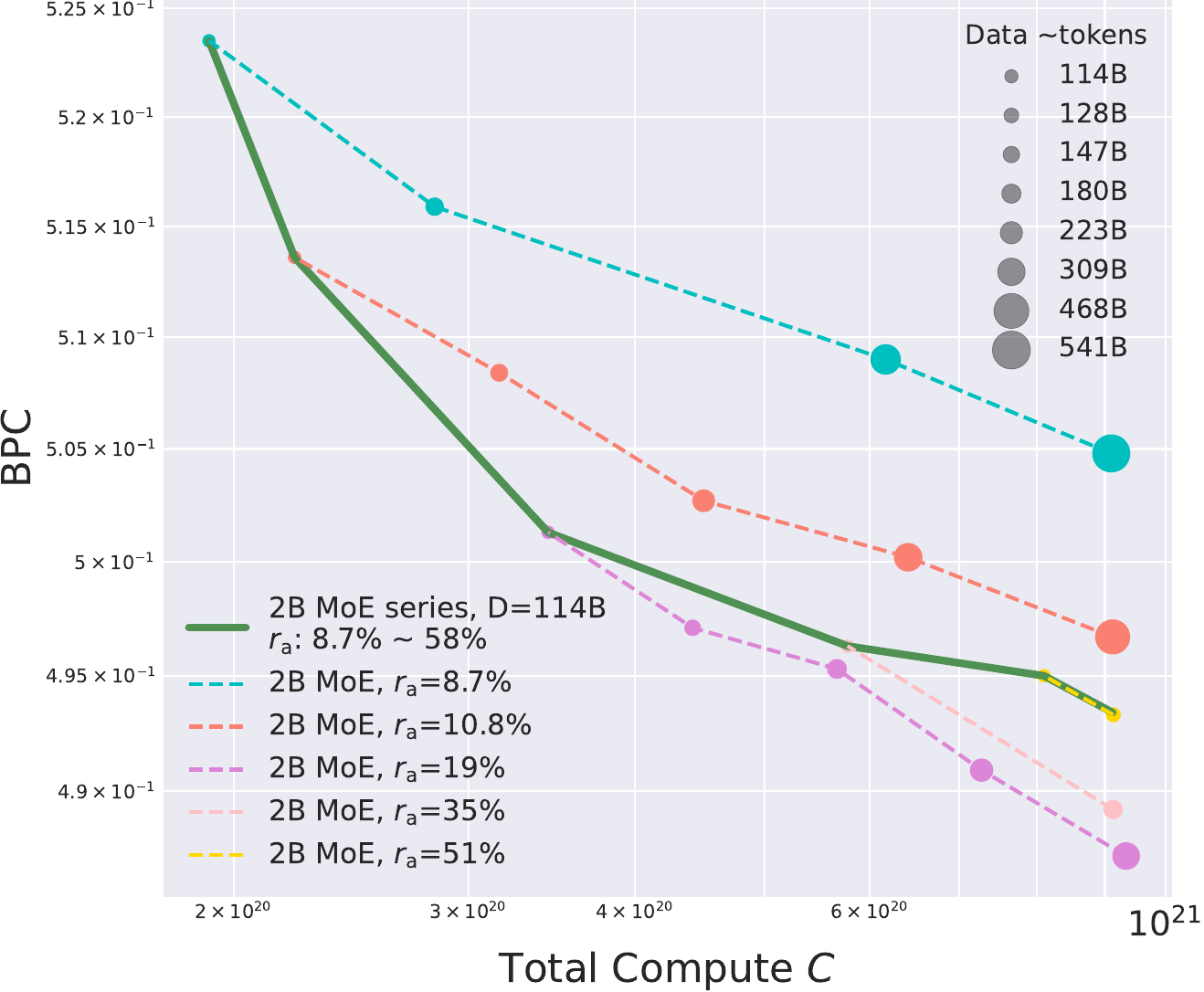}
        \caption{Fixed $D$ (solid) or $r_\text{a}$ (dashed)}
        \label{fig:2B_optimal_ra_keepD}
    \end{subfigure}
    \hspace{3mm}
    \begin{subfigure}{0.4\linewidth}
        \includegraphics[width=\linewidth]{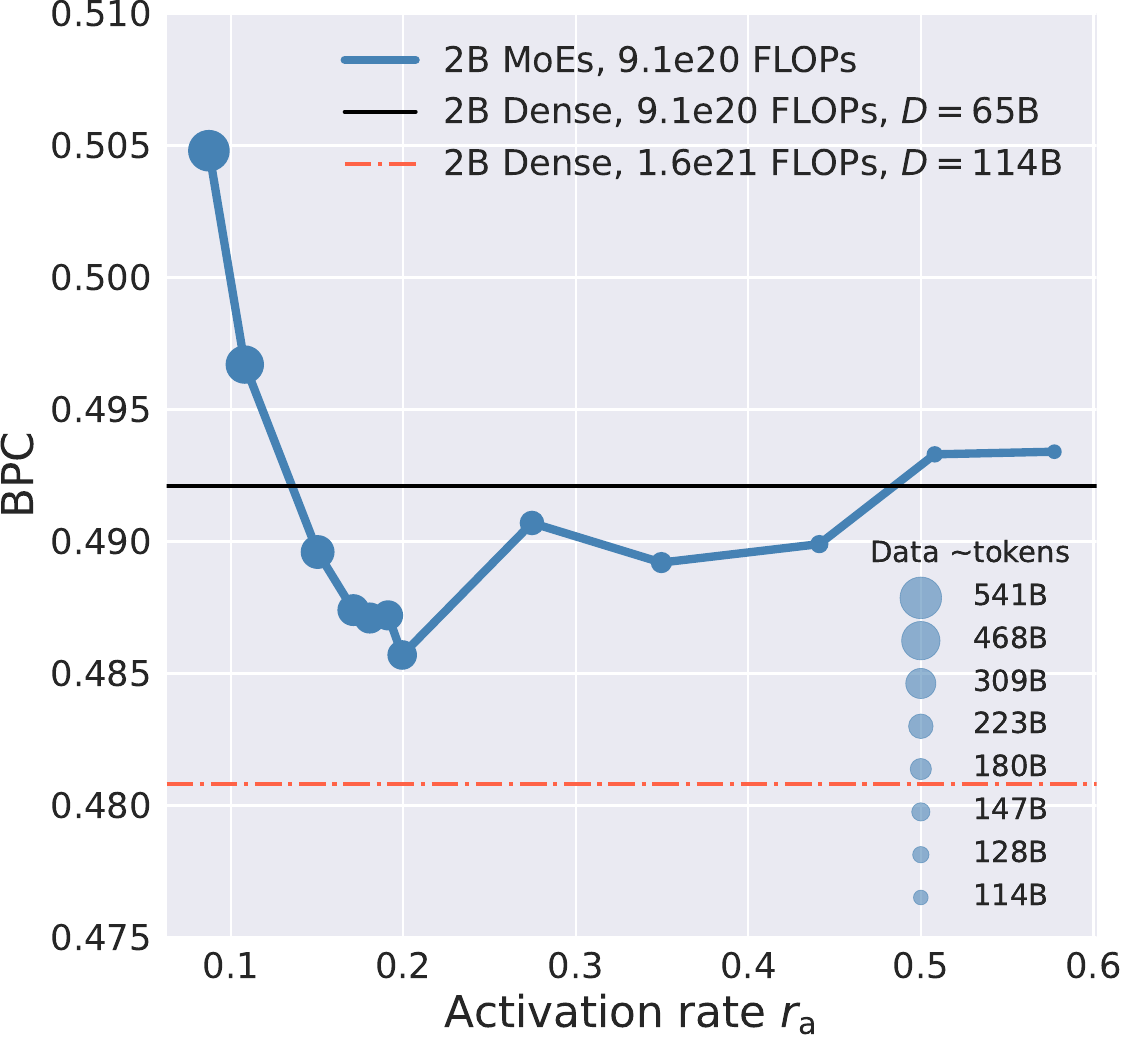}
        \caption{Fixed $C$}
        \label{fig:2B_optimal_ra_keepC}
    \end{subfigure}
    \caption{        
        Performance of $N \approx 2\text{B}$ models trained with varying data sizes $D$ and activation rates $r_\text{a}$.
        \textbf{(a)} With a fixed $D$, performance gain exhibits a \textbf{non-linear} dependence on training budget $C$. 
        Conversely, with a fixed $r_\text{a}$, increasing $D$ leads to a \textbf{linear} performance gain. 
        These findings indicate an optimal activation rate, $r_\text{a}^{**}$ = 20\%, that is consistent across various $D$ values when $N$ is constant.
        \textbf{(b)} With a fixed training compute $C$, the optimal activation rate $r_\text{a}^{**}$ = 20\% can be clearly seen.
        % We conclude three key findings for MoE models: 
        % 1) BPC decreases with increasing $D$.
        % 2) The performance gain does not depend linearly on the compute budget.
        % 3) An MoE model whose $r_\text{a}$ resides in the outstanding region $R_\text{a}^*$ is able to outperform its dense counterpart given the same compute $C$.
    }
    \label{fig:2B_optimal_ra}
\end{figure*}

\section{Optimized MoE Architecture}
\label{sec:arc}

%The model backbones for exploring various ARs are based on careful designs and thorough considerations to ensure the fairness of our comparisons, and more importantly, to discover the optimal ARs.
%Therefore, we will first introduce these model configurations in this section before diving deep into the discussion of optimal ARs.
%We conduct experiments for different components in MoE models to optimize the model architecture, and the conclusions are applied on our backbone design in \S~\ref{sec:activation_rate}.
%We detail each of our explorations below.
% We summarize our explorations and results in Table~\ref{} and will detail each of them below.
Building on the insights discussed above, we systematically examine the following model components in the order outlined below:
\begin{inparaenum}[1)]
    \item Distribution of MoE and dense layers.
    \item Gate score normalization.
    \item Parameter allocation within MoE.
    \item Exploration of optimal structural hyperparameters.
\end{inparaenum}
Each component incorporates previous conclusions into its experimental settings.

\noindent\textbf{MoE and dense layers arrangement.}
This part examines how to arrange the distribution between MoE and dense layers.
We consider three layer arrangement schemes: every layer is an MoE layer (\texttt{full}), one dense layer followed by MoE layers (\texttt{1dense}), and interleaved MoE and dense layers (\texttt{interleave}).
% In the \texttt{full} scheme, all layers are MoE layers.
% The \texttt{1dense} scheme employs a dense layer as the first layer and MoE layers for the remaining layers, while \texttt{interleave} interleaves dense layers with MoE layers in an alternating pattern.
We additionally include shared experts (SE) for some of our experiments.

Table~\ref{tab:layer_arrangement} in Appendix~\ref{appedix-results} presents the experimental settings and results.
The conclusions are as follows: 
\begin{inparaenum}[1)]
    \item \texttt{1dense}+SE performs the best, possibly because the dense layer contributes to more stable training.
    \item The ratio of shared expert size to total expert size $\sfrac{D_\text{se}}{(D_\text{se}+KD_\text{e})}$ has minimal impact on model performance.
\end{inparaenum}
Therefore, we continue using the \texttt{1dense}+SE configuration and set $D_\text{se}=KD_\text{e}$.

\noindent\textbf{Gate score normalization.}
The results of normalizing gate scores of chosen experts are in Table~\ref{tab:score_norm} in Appendix~\ref{appedix-results}.
% \begin{equation}
%     g_i(x) = 
%     \begin{cases}
%          \dfrac{s_i}{\sum_i g_i} & \text{if } s_i \in \operatorname{TopK}(s; K), \\
%          0 & \text{otherwise}, 
%     \end{cases} \\
% \end{equation}
Although the addition of normalization does not show an obvious difference in performance loss, it tends to reduce the  average balance loss $\mathcal{\bar{L}_\text{balance}}$.
% ~\footnote{We adopt the expert-level balance loss proposed in \citep{dai2024deepseekmoe}}.
Since normalization requires $K$ > 1 to avoid zero gradient, we opt not to normalize given that some of our experiments have $K$=1.

\noindent\textbf{Top-K setting.}
In this part, we discuss the allocation of parameters within MoE layers, focusing on the top-$K$ setting.
Expert granularity is adjusted by varying $K$ and $D_\text{e}$ while keeping their product constant. 
%, i.e., $K \cdot D_\text{e} = \text{constant}$.
We conduct three groups of experiments with various $r_\text{a}$ and the results are given in Table~\ref{tab:fine_grained} in Appendix~\ref{appedix-results}.
Note that within each experiment group, the product $K \cdot D_e$ is not strictly maintained due to compatibility with other hyperparameters. 
We observe that both overly large $K$ and the $K$=1 setting are generally suboptimal across the three groups.
Therefore, we avoid using large $K$ and avoid setting $K$=1 in our main experiments whenever possible.

% As discussed in the previous section, the optimal activation rate region possibly enables stronger expert specialization, and thus activating more specialized experts is likely to achieve better results in these circumstances. 
% We adopt small $D_\text{e}$ for experiments in \S~\ref{sec:activation_rate} whenever possible.

\noindent\textbf{Model shape ratios.}
As discussed in \S~\ref{subsec:method_consideration}, the shape hyperparameters include three ratios: $\zeta$, $\alpha$, and $\beta$.
We set $\alpha=2.77$~\citep{touvron2023llama2} and explore the optimal $\zeta$ and $\mu$, from which $\beta$ can be derived.
% The two ratios related to model shape are $\zeta=\frac{D_\text{m}}{L}$ and $\mu=\frac{D_\text{se}+ED_\text{e}}{D_\text{m}}$.
As illustrated in Fig.~\ref{fig:hidden_dimention} in Appendix~\ref{appedix-results}, although performance fluctuates wildly given a value of $\zeta$ or $\mu$, there is an overall upward trend with increasing $D_\text{m}$ for $\zeta$ and a downward trend for $\mu$.
Following the observed trend, we set $\zeta \approx 88$ and $\mu \approx 22$ for the subsequent experiments.

% whether specific relationship between model hidden dimension $D_\text{m}$ and other hyperparameters exists.
% We experiment with $D_\text{m}$ ranging from 12 to 49.
% More experimental settings can be found in Table~\ref{}.
% Although the current number of experiments is not enough to derive confident conclusions, we still observe several tendencies.
% When $D_\text{m}$ increases, we have (Figure~\ref{}):
% \begin{inparaenum}[1)]
%     \item The minimum $\mathcal{L}$ becomes lower.
%     \item The optimal value of $L$, $\mu$, and $\beta$ becomes smaller.
%     \item The optimal ratio of self-attention parameters to the total model parameters becomes larger. 
% \end{inparaenum}
% Given that $D_\text{m}$ and $L$ have a huge influence on the model performance, we keep them unchanged within each group of experiments in \S~\ref{sec:activation_rate} to control variables and focus on $r_\text{a}$~\footnote{The optimal $D_\text{m}$ and $L$ for each $r_\text{a}$ might differ and, ideally, we should compare the performance of each $r_\text{a}$ where $D_\text{m}$ and $L$ are set to be the corresponding optimal values rather than fixed values.}.

% \begin{table}[ht]
%     \centering
%     \small
%     \begin{tabular}{ccccc}
%     \toprule
%     $\lambda$       & $\xi$       & $\omega$      & $c$              & $d$              \\ 
%     \midrule
%     -0.713    & 0.307& 0.571& 1.79 & 0.58 \\ 
%     \bottomrule
%     \end{tabular}
%     \label{tab:fit_coeff}
% \end{table}
\begin{figure*}[t!]
    \centering
    \begin{subfigure}{0.45\linewidth}
        \includegraphics[width=\linewidth]{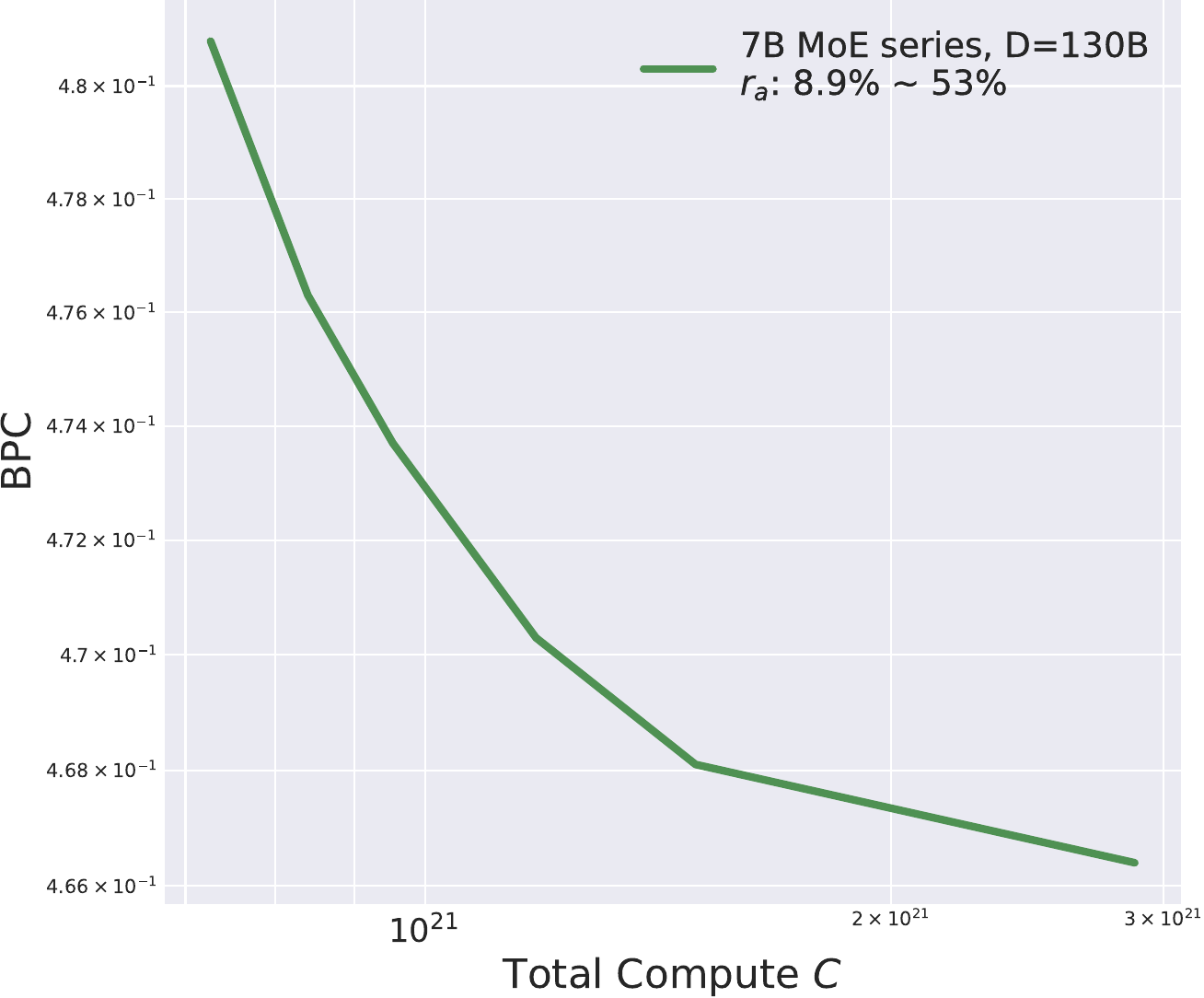}
        \caption{Fixed $D$}
        \label{fig:7B_optimal_ra_keepD}
    \end{subfigure}
    \hspace{3mm}
    \begin{subfigure}{0.40\linewidth}
        \includegraphics[width=\linewidth]{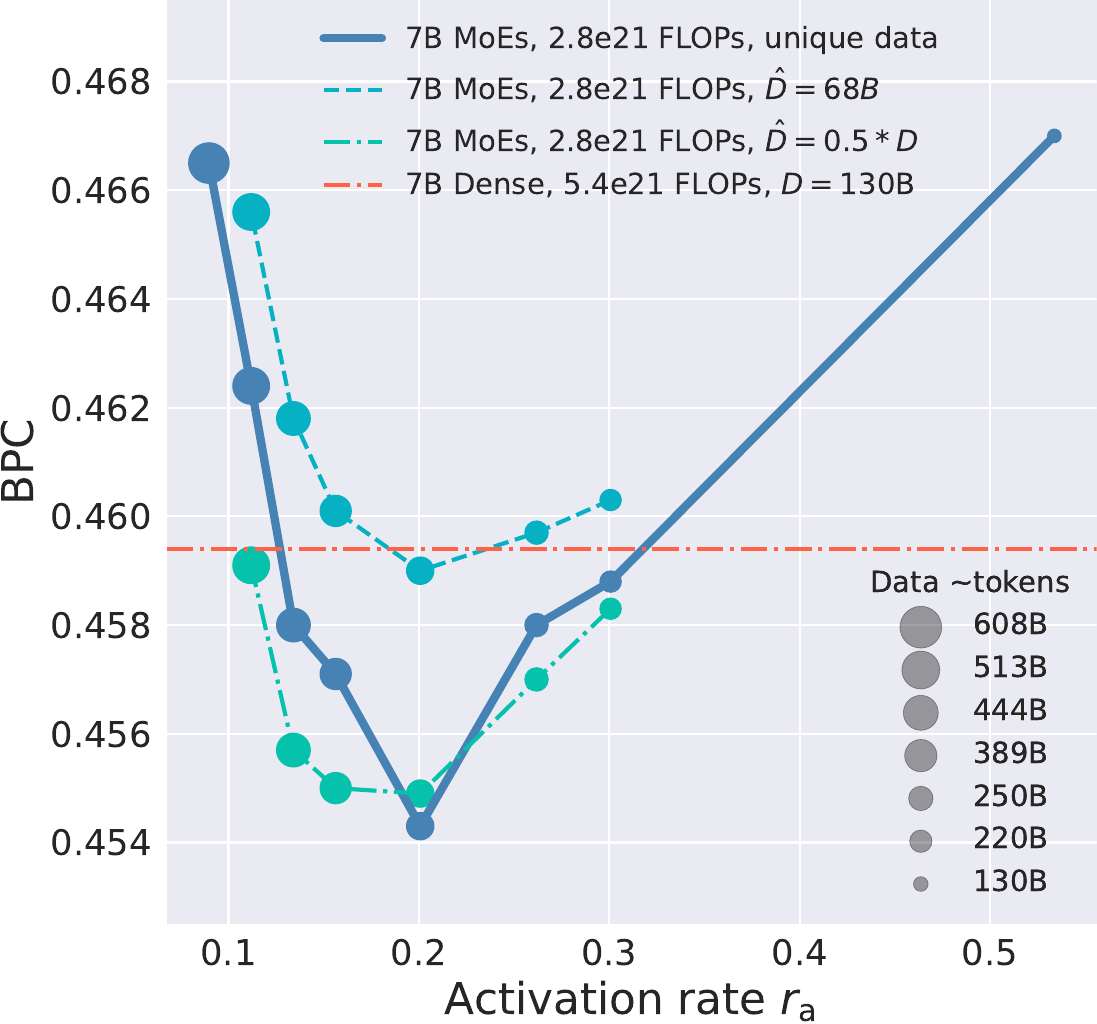}
        \caption{Fixed $C$ and reusing data}
        \label{fig:7B_optimal_ra_keepC}
    \end{subfigure}
    \caption{
        Performance of $N \approx 7\text{B}$ models trained with varying data sizes $D$ and activation rate $r_\text{a}$.
        The optimal activation rate, $r_\text{a}^{**}=20\%$, align with the findings for the 2B models (Figure~\ref{fig:2B_optimal_ra}).
        Additionally, compared to training on the unique dataset, the strict data reuse scheme shows only a slight performance reduction, while the loose scheme often yields better performance.
    }
    \label{fig:7B_optimal_ra}
\end{figure*}

\section{Optimal Activation Rate}
\label{sec:activation_rate}

% Existing works find that the performance of MoE models tends to be inferior to their dense counterparts with the same compute budget.
% However, this observation might be an incomplete picture due to the lack of comprehensive and systemic studies.
In this section, we analyze how the performance of MoE LLMs varies with different activation rates (AR) using model backbones optimized based on the conclusions in \S~\ref{sec:arc}, and examine whether MoE models can outperform dense models. Note that a concurrent study~\citep{abnar2025parameters} suggests that the optimal sparsity of MoE depends on model capacity.
However, our findings indicate that, with optimized backbones, the optimal AR remains \textbf{consistent} across models of different sizes.
We first detail our experimental setup and results, followed by a further discussion on the conclusions.
% Then we study the feasibility of data reusing to address the additional data requirement of MoE trainings.
% Finally, we investigate the downstream performance of our trained models. 

% In this section, we introduce the details of our experiments, which focus on optimal sparsity of MoE LLMs, data reusability, MoE model architecture, and optimal hyperparameter estimation, respectively. 
% For each experiment, we present its goal, experimental settings, results, and discussion. 

\noindent\textbf{Setup.} 
We built a series of MoE models with non-vocabulary parameters $N \approx 2\text{B}$ and $N \approx 7\text{B}$, but varying activation rates $r_\text{a}$ from 8.7\% to 58\%.
Noteworthy, the model backbones were built upon the findings in \S~\ref{sec:arc}, as detailed in Table~\ref{tab:2B_keepD}, \ref{tab:2B_moe_ra}, \ref{tab:7B_keepD} and \ref{tab:7B_moe_ra} in Appendix~\ref{appedix-results}.
% In \S~\ref{sec:arc}, we have explored the optimization of MoE architecture.
% Building upon the aforementioned findings, we construct a series of MoE models with nearly identical number of non-vocabulary parameters $N \approx 2\text{B}$ and $N \approx 7\text{B}$, but with different activation rates $r_\text{a}$ ranging from 8.7\% to 58\%.
% The experimental settings are summarized in Table~\ref{tab:2B_keepD},\ref{tab:2B_moe_ra},\ref{tab:7B_keepD},\ref{tab:7B_moe_ra}.
Each model was trained on a proportional subset of our dataset, ensuring $\sfrac{D}{N} \geq 20$~\citep{hoffmann2022training} for sufficient training.

\subsection{Optimal AR Point}

Focusing on the 2B models trained on the same data size $D=114\text{B}$ as shown by the green solid line in Figure~\ref{fig:2B_optimal_ra_keepD}, we observe that the performance gain \textbf{depends non-linearly} on the training budget $C$. 
Specifically, the gain is more significant within a relatively low range of $r_\text{a}$.
Starting from points on this curve and fixing the corresponding $r_\text{a}$ values, increasing $D$ results in \textbf{linearly diminishing} BPC, as indicated by the dashed lines in Figure~\ref{fig:2B_optimal_ra_keepD}.
% ~\footnote{The overall performance at $r_\text{a}=8.7\%$ (blue dashed line) could be improved since we set $K=1$ for this set of experiments, an unfavorable configuration as demonstrated in \S~\ref{sec:arc}.}.
These results confirm the \textbf{existence of an optimal AR point, $r_\text{a}^{**}$, that remains consistent regardless of $D$ when $N$ is unchanged}. 
When plotting the results from a fixed training compute perspective ($C=C_0=9.1\mathrm{e}20$) in Figure~\ref{fig:2B_optimal_ra_keepC}, we clearly observe that the optimal AR point is approximately $r_\text{a}^{**} \approx 20\%$.

\subsection{Comparison with Dense Models}
\label{subsec:comparsion_dense}

To compare with dense models, we trained two dense models (Table~\ref{tab:dense_baseline} in Appendix~\ref{appedix-results}) with $N \approx 2\text{B}$ parameters and training budgets $C_1=C_0=9.1\mathrm{e}{20}$ and $C_2=1.64\mathrm{e}{21}\approx2C_1$. 
The second model ($C_2$) is included for comparison to account for the typically lower Model FLOPs Utilization (MFU) in MoE training. 
This reduced MFU arises from load balancing and expert parallelism mechanisms that limit large-block matrix computations.
As illustrated in Figure~\ref{fig:2B_optimal_ra_keepC}, MoE models outperform their $C_1$ dense counterparts when $r_\text{a}$ falls within a specific range (approximately 15\% to 48\% for 2B models). 
For instance, the MoE model with the optimal AR point $r_\text{a}^{**}=20\%$ achieves a BPC value that is 0.0064 lower than its $C_1$ dense counterpart and only 0.0049 higher than the $C_2$ dense model. 
This demonstrates the existence of an \textit{optimal activation rate region} $R_\text{a}^*$, where \textbf{MoE models with $r_\text{a} \in R_\text{a}^*$ can outperform their dense counterparts under the same training budget $C$} and approach the performance of dense models with double the compute.
However, the performance gains of MoE models rely on a substantial increase in data, \eg, a $4.6\times$ larger data size at $r_\text{a}=r_\text{a}^{**}=20\%$. 
To mitigate this additional data requirement, we explore a data reuse strategy in \S~\ref{sec:data_reuse}.

\begin{figure*}[t!]
    \centering
    \includegraphics[width=\linewidth]{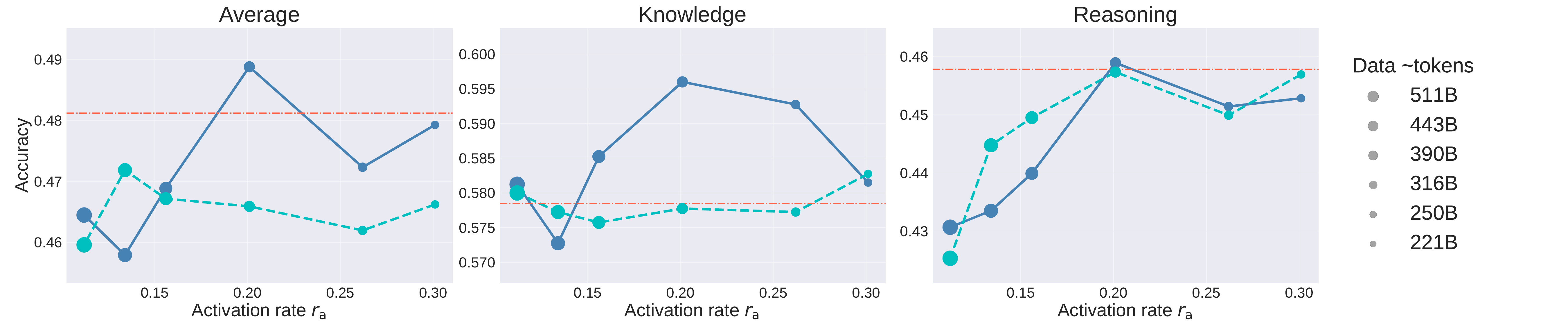} \\
    \includegraphics[width=\linewidth]{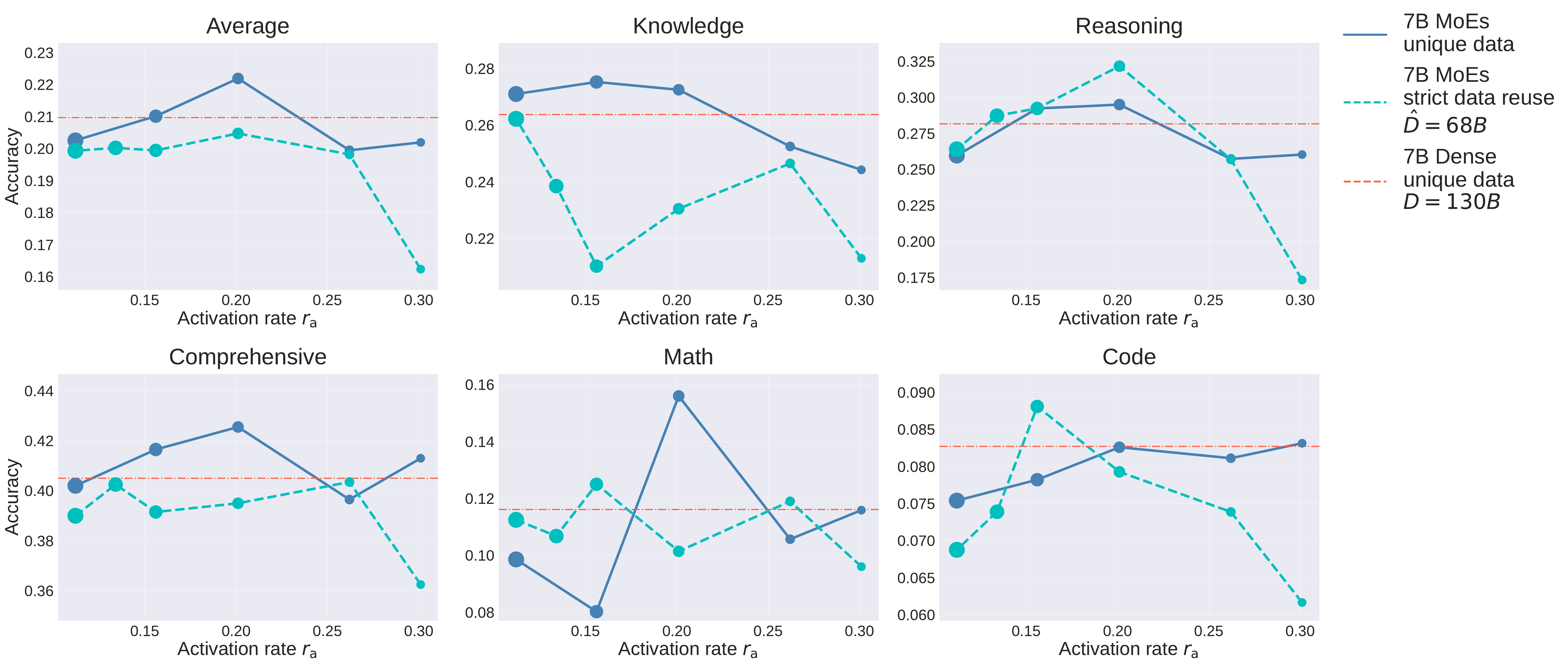}
    \caption{
        Downstream performance of 7B models: pre-trained (\textbf{top}) and SFT-ed (\textbf{middle and bottom}) versions. 
        Across all benchmark types, MoE models with $r_\text{a}=20\%$ outperform dense model trained with twice the compute, aligning with upstream observations that the optimal AR is 20\%.
    }
    \label{fig:downstream}
\end{figure*}

\subsection{Consistency of Optimal AR}

As illustrated in Fig.~\ref{fig:7B_optimal_ra}, an optimal AR point $r_\text{a}^{**}$ also exists for 7B models. 
Surprisingly, $r_\text{a}^{**}$ remains consistent for both 2B and 7B models at approximately 20\%, suggesting that \textbf{$r_\text{a}^{**}$ is independent of model size}. 
This finding contradicts established studies on MoE sparsity~\citep{abnar2025parameters}, which proposes that optimal sparsity (defined as $\sfrac{(E-K)}{E}$) is directly proportional to model size. 
Nevertheless, our experiments were conducted with strictly controlled variables using optimized backbones, leading us to believe that our conclusions are both \textbf{reliable} and \textbf{scalable} (see Appendix~\ref{append:related_work} for a more detailed discussion). 
To further validate the possible universality of our findings, we conducted experiments on $N \approx 3\text{B}$ models and achieved similar results (see Fig.~\ref{fig:3B_optimal_ra} in Appendix~\ref{appedix-results}).

Expert specialization is another significant potential of MoEs in addition to remarkable scalability, where each expert focuses on learning specific features or patterns within the data. 
However, this attribute has not yet been clearly observed even in state-of-the-art MoE LLMs~\citep{lo2024closer,zhang2024diversifying}, and effective approaches to achieve it remain under-explored. 
Based on our observation that MoEs outperform dense models when $r_\text{a} \in R_\text{a}^*$, we conjecture a relationship between the optimal AR region and the degree of expert specialization. 
Specifically:
\begin{inparaenum}[1)]
    \item When the activation rate is too low ($r_\text{a} < 10\%$), the model lacks sufficient parameters to store knowledge effectively.
    \item When the activation rate is relatively high ($r_\text{a} > 50\%$), more experts are typically activated, which may lead to weaker specialization.
\end{inparaenum}
An activation rate within the optimal region $R_\text{a}^*$ likely facilitates a higher degree of expert specialization, thereby enhancing the MoE model's performance compared to its dense counterpart. 
We leave further analysis of this potential relationship for future work.

% Due to computational cost, the 7B dense model on $C=2.8e21$ are trained with $\sfrac{D}{N}<20$ and thus might suffer from under-training, leading to unsatisfying results.
% Directly comparing the performance of MoEs with theirs are unfair.
% As a result, we are unable to determine the outstanding activation region $R_\text{a}^*$ for 7B models so far.
% We evaluate the MoE models on several benchmarks to examine their downstream performance.
% As illustrated in Figure~\ref{}, the performance of MoE models on downstream tasks follows a similar trend to the BPC.
% Moreover, , but in a different model family~\citep{} with $\sfrac{D_\textbf{m}}{L}=58$, while the 2B and 7B models have $\sfrac{D_\textbf{m}}{L} \approx 88$. 
% Table~\ref{} depicts the model settings.
% Again, we observe $r_\text{a}^{**}=19$ in Figure~\ref{}~\footnote{Given that the curves of reusing data or not are alike, we }, which shows great promise that this optimal activation rate are \textbf{robust across model families}.
\begin{table}[t!]
    \centering
    \footnotesize
    \smaller[0.5]
    \caption{Accuracy of 7B SFT-ed models across different benchmarks.}
    \label{tab:bmk}
    \begin{tabular}{llccc}
        \toprule
        & & Dense baseline & \multicolumn{2}{c}{MoE w/ optimal AR} \\
        \midrule
        \multirow{3}{*}{Pretrain info} & Activation rate & - & 20.07 & 20.07 \\
        & Compute & 5.45e21 & 2.86e21 & 2.86e21 \\
        & Data reuse & - & - & strict \\
        \midrule
        \multirow{4}{*}{Knowledge} & CMMLU~\citep{li2023cmmlu} & 31.23 & 31.62 & \textbf{32.11} \\
        & MMLU~\citep{hendrycks2020measuring} & 31.26 & \textbf{32.92} & 24.57 \\
        & MMLU-Redux~\citep{gema2024we} & 28.90 & \textbf{30.93} & 23.73 \\
        & MMLU-Pro~\citep{wang2024mmlu} & \textbf{14.12} & 13.59 & 13.59 \\
        \midrule
        \multirow{3}{*}{Reasoning} & DROP~\citep{dua2019drop} & 32.32 & \textbf{35.13} & 30.93 \\
        & LiveBench~\citep{white2024livebench} & 16.82 & \textbf{18.15} & 16.76 \\
        & MUSR~\citep{sprague2023musr} & 35.98 & 35.58 & \textbf{48.94} \\
        \midrule
        \multirow{2}{*}{Comprehensive} & AGIEval~\citep{zhong2023agieval} & 20.89 & \textbf{22.07} & 21.02 \\
        & BBH~\citep{suzgun2022challenging} & 58.02 & \textbf{60.01} & 56.07 \\
        \midrule
        \multirow{2}{*}{Math} & GAOKAO-Math24~\citep{zhang2023evaluating} & 9.92 & \textbf{15.70} & 9.09 \\
        & GSM8K~\citep{cobbe2021training} & 13.34 & \textbf{15.54} & 11.22 \\
        \midrule
        \multirow{5}{*}{Code} & APPS~\citep{hendrycks2021measuring} & 7.35 & 6.80 & \textbf{8.18} \\
        & DS-1000~\citep{lai2023ds} & 5.70 & \textbf{6.90} & 4.60 \\
        & HumanEval~\citep{chen2021evaluating} & \textbf{22.56} & 21.34 & 21.95 \\
        & LeetCode~\citep{coignion2024performance} & 1.49 & \textbf{1.67} & 1.49 \\
        & LiveCodeBench~\citep{jain2024livecodebench} & 4.21 & \textbf{4.63} & 3.37 \\
        \bottomrule
    \end{tabular}
\end{table}

\section{Data Reuse Strategy}
\label{sec:data_reuse}

As discussed in \S~\ref{subsec:comparsion_dense}, MoEs outperform their dense counterparts but require additional data. 
To eliminate this increased data demand, we investigate data reusability by training models for multiple epochs using a fixed, smaller dataset size $\hat{D}$. 
We extract a sub-dataset of size $\hat{D}$ from the original training dataset. 
At the beginning of each epoch after the first, the data are shuffled.

\noindent\textbf{Setup.}
We explore two distinct schemes, termed the \textit{strict} and the \textit{loose} data reuse schemes.

For the strict scheme, our aim is to ensure that both MoE and dense models are trained under \textit{completely equal conditions with respect to $N$, $D$, and $C$}.
Given a fixed $\hat{D}$, the number of training epochs (ranging from 1.7 to 8.3 in our 3B and 7B model experiments) increases as $r_\text{a}$ decreases (hence decreasing $M$) to maintain the compute budget $C$.
The experimental settings are detailed in Table~\ref{tab:7B_data_reuse},~\ref{tab:3B_reuse_65B},~\ref{tab:3B_reuse_114B}, in Appendix~\ref{appedix-results}, which are mostly the same as those in \S~\ref{subsec:comparsion_dense}, except for the training data used. 
Specifically, we set $\hat{D} = 65\text{B}$ and $114\text{B}$ for the 3B models, and $\hat{D} = 68\text{B}$ for the 7B models, corresponding to the data used for training the dense models.

For the loose scheme, we relax the constraint of identical $D$ by fixing the number of training epochs to 2 for all $r_\text{a}$, hence $\hat{D}=0.5D$, where the exact value of $D$ corresponds to the specific $r_\text{a}$.
We conduct experiments on 7B models and the experimental settings are in Table~\ref{tab:7B_loose_reuse} in Appendix~\ref{appedix-results}.

% To maintain the compute budget $C$, the number of training epochs increases as the activation rate decreases (and thus $M$ decreases), ranging from 1.7 to 8.3 epochs in our experiments. 
% This approach ensures that both MoE and dense models are trained under \textit{completely equal conditions} regarding $N$, $D$, and $C$.

\noindent\textbf{Results.}
The performance under the strict scheme is illustrated by the blue dashed lines in Fig.~\ref{fig:7B_optimal_ra_keepC} and ~\ref{fig:3B_optimal_ra} in Appendix~\ref{appedix-results}.
Reusing data $\hat{D}$ only marginally diminishes performance compared to training on the unique dataset $D$ for a single epoch, and MoE models continue to outperform dense baselines. 
Moreover, increasing $\hat{D}$ further narrows the performance gap. 
The similarity in curve shapes indicates that the optimal activation rate $r_\text{a}^{**}$ remains unchanged. 
These findings address the primary question posed at the beginning of this paper: 
\textbf{Mixture-of-Experts can surpass dense LLMs under equal total parameters, compute, and data constraints, provided that the backbones are optimized and $r_\text{a} \in R_\text{a}^*$}. 
We further discuss the reuse-vs-unique trade-offs below.

\noindent\textbf{Discussion.} 
Prior works have explored the effectiveness of multi-epoch training for dense and MoE models.
\citet{muennighoff2023scaling} developed a scaling law that accounts for the number of repeated tokens and found negligible loss for repeating up to 4 epochs compared to training on unique data, whereas \citet{hernandez2022scalinglawsinterpretabilitylearning} showed degradation for dense models.
\citet{xue2023repeatrepeatinsightsscaling} noticed no significant gain for MoEs with repeated training when high-quality data is insufficient. 
We emphasize that our goal here is \emph{not} to claim that multi-epoch training is generally better for MoEs; 
rather, we examine whether MoEs can still surpass dense models when the \emph{unique-token budget} is fixed.
Concretely, under the loose scheme (green dashed line), for each $r_\text{a}$ we keep the consumed-token budget $D$ fixed and compare 
(i) a 2-epoch run on a subset of size $\hat{D}=0.5D$ (thus processing $D$ tokens with reuse) and 
(ii) a 1-epoch run that consumes $D$ tokens without reuse (i.e., $D$ unique tokens), where both are sampled from the same data recipe/distribution. 
We find that the 2-epoch reuse setting can match, and sometimes slightly improve over, the 1-epoch unique-token baseline at several suboptimal $r_\text{a}$ points;
however, for the 7B models, at the most important optimal point ($r_\text{a}\approx 20\%$) it does not exceed the 1-epoch unique-token baseline.
In all case for the 7B models, using more than two epochs (multi-epoch) consistently degrades performance.
For the 3B models under the strict reuse setting, at a fixed $r_\text{a}$, using a larger unique-token budget (e.g., $\hat{D}=114\text{B}$) 
consistently outperforms a smaller one (e.g., $\hat{D}=65\text{B}$), 
aligning with the intuition that more unique data is better even under multi-epoch training 
(see Fig.~\ref{fig:3B_optimal_ra} and Tables~\ref{tab:3B_reuse_65B}--\ref{tab:3B_reuse_114B} in Appendix~\ref{appedix-results}). 
Moreover, increasing the unique-token budget from 114B (trained for 2--3 epochs) to 309B (trained for 1 epoch) yields only a marginal improvement under the same token-consumption budget (Fig.~\ref{fig:3B_optimal_ra}).
Moreover, under a fixed consumed-token budget ($D=309$B), increasing the unique-token budget from $\hat{D}=114$B (trained for $\sim$2--3 epochs with reuse) to 309B (trained for 1 epoch without reuse) yields only a marginal improvement (Fig.~\ref{fig:3B_optimal_ra}).
Overall, these results suggest that MoE models may tolerate mild repetition (around two epochs) under a fixed token-consumption budget, but additional repetition becomes harmful.
% We would like to emphasize the huge practical value of this conclusion: with exactly the same data size $D$ and training budget $C$, we can achieve a MoE model that is much more capable and efficient (\eg, 1\% lower BPC and 5 $\times$ lower $\hat{C}$ for the 2B model whose $r_\text{a}=r_\text{a}^{**}$) than its dense equivalent.
\section{Analysis of Downstream Performance}
\label{sec:downstream}

To assess whether the optimal ARs generalize to downstream tasks, we conduct SFT on our 7B pre-trained models (trained w/ and w/o strict data reuse) and evaluate both the pre-trained models and SFT-ed models on a total number of 29 benchmarks (see Fig.~\ref{fig:downstream} and Tab.~\ref{tab:bmk}), including categories such as \texttt{reasoning} and \texttt{knowledge}.
The comprehensive list of benchmarks can be found in Appendix~\ref{append:bmk}.
% ~\footnote{While the sequence length of our models is limited to $S=2048$, resulting in suboptimal absolute accuracy, we can still analyze their relative performance.}.  
For all SFT trainings, we use a fixed data size $D$, and thus varying $C$ across models with different $r_\text{a}$.
The 7B dense model trained with \textit{twice} the compute is included for comparison. 

\noindent\textbf{MoE \textit{vs.} Dense at $r_\text{a}=r_\text{a}^{**}$.}
For both PT and SFT models, MoEs outperform their dense equivalents across all benchmark types when $r_\text{a} = 20\%$. 
This result aligns with upstream findings that the optimal activation rate ($r_\text{a}^{**}$) is 20\%, highlighting the very possible \textit{universality of the optimal AR point across different training phases and data domains}.
Furthermore, $r_\text{a}^{**}$ remains unchanged during SFT, even with varying $C$, suggesting that the SFT data size may have an upper limit for performance improvement, provided the PT model is adequately trained. 
Additionally, the \texttt{average} performance at $r_\text{a} \neq r_\text{a}^{**}$ is consistently inferior to that of dense models, underscoring the critical role of the optimal AR point.

\noindent\textbf{MoE \textit{vs.} Dense at $r_\text{a}<r_\text{a}^{**}$.}
Dense models outperform MoEs across all domains for PT models. 
After SFT, MoEs overtake dense models on \texttt{comprehensive} and \texttt{knowledge} tasks. 
However, a notable performance gap remains in \texttt{math}, highlighting the PT stage's importance for math ability.

\noindent\textbf{MoE \textit{vs.} Dense at $r_\text{a}>r_\text{a}^{**}$.}
Compared to the dense models, MoEs perform better on \texttt{knowledge} but worse on \texttt{reasoning} for PT models, and usually slightly underperform after SFT.

\noindent\textbf{Sparser \textit{vs.} Denser.}
For PT models, denser MoEs (\(r_\text{a} > r_\text{a}^{**}\)) outperform or match the performance of sparser MoEs (\(r_\text{a} < r_\text{a}^{**}\)) across all domains, consistent with Figure~\ref{fig:7B_optimal_ra_keepC}, regardless of data reuse.
When training on unique data, sparser MoEs perform better on \texttt{knowledge}. 
Notably, denser MoE performance significantly degrades with data reuse, especially for SFT.

\noindent\textbf{Impact of data reuse.}
For both PT and SFT models, data reuse has little impact on \texttt{reasoning} but causes significant degradation in \texttt{knowledge} performance. 
Surprisingly, at \(r_\text{a} = r_\text{a}^{**}\), the SFT-ed MoEs trained with data reuse outperform both MoEs and dense models trained on unique data. 
This implies that \textit{a model can master reasoning skills (rather than merely memorizing information~\citep{DBLP:conf/icml/HuTYZ24}) with a relatively small dataset~\citep{muennighoff2025s1,wang2025reinforcement} and further enhance its capabilities through multiple training epochs}.
\section{Conclusion and Future Works}
\label{sec:conclusion}

In this paper, we propose a three-step experimental methodology to investigate whether MoEs can surpass their dense counterparts under the same constraints on total parameters, compute, and data. 
By optimizing the architecture, identifying the optimal activation rate region, and reusing data, we arrive at a positive answer to this question. 
Future work will explore how optimal activation rates enhance model capabilities and whether similar conclusions hold for other training methods like upcycling~\citep{komatsuzaki2022sparse} and MoEfication~\citep{zhang2021moefication}.
We hope this work offers valuable insights for the architectural design of next-generation models.
% we demonstrate that given a fixed computational budget, MoE Transformers do not necessarily underperform their dense equivalents and can be superior with optimized architectural designs.
% One of the key factors is the activation rate, whose values exist an optimal region that is independent of model sizes.
% By reusing data, the extra requirement of data to achieve this performance gain for MoEs can be fully addressed. 

% We speculate that a suitable activation rate may promote expert specialization, leading to improved performance. 
% Since our experiments focus solely on training MoEs from scratch, another avenue for future research will be to investigate whether similar conclusions hold for other training approaches, such as upcycling~\citep{komatsuzaki2022sparse} and MoEfication~\citep{zhang2021moefication}.

\section*{Limitations}

The limitations of this work include:
\begin{inparaenum}[1)]
    \item Hindering by the high computational cost, we did not train models larger than 7B.
    \item As described in \S~\ref{sec:arc}, we focus mainly on the impact of several main components of MoEs, but fix the rest to narrow the scale of experiments.
    \item Exploration of other elements can provide further comprehensive guidance for the architectural design of MoEs.
% However, we believe that our conclusions are scalable given that the optimal activation rate region remains consistent when scaling the model size from 2B to 7B.
% Moreover, our experimental results are limited to pretraining and did not investigate the applicability for downstream tasks. 
\end{inparaenum}

\bibliography{iclr2026_conference}
\bibliographystyle{iclr2026_conference}

\appendix
% \section{Appendix}
\section*{Appendix}

\section{Background: Mixture-of-Experts}
\label{append:background}

The MoE architecture primarily consists of a gate and several experts.
Typically, the gate $g(\cdot)$ is composed of a linear layer $W_g$ followed by a Softmax and a Top-K operation, and the experts $\{E_i\}_{i=1}^{E}$ follow the standard FFN structure.
The computation of an MoE block can then be formulated as follows:
\begin{align}
    s_i(x) &= \operatorname{Softmax}_i(W_\text{g} x), \\
    \mathcal{T}(x) &= \operatorname{TopK}(s(x); K), \\
    g_i(x) &= 
    \begin{cases}
         s_i(x) & \text{if } i \in \mathcal{T}(x), \\
         0 & \text{otherwise}, 
    \end{cases} \\
    y &= \sum_{i=1}^{E} g_i(x) \cdot E_i(x),
\end{align}
where $s_i(x)$ denotes the gate score for the $i$-th expert. Unless otherwise specified, we use the above \emph{non-normalized} Top-$K$ gating in this paper (\ie, we do not renormalize the $K$ selected scores to sum to 1). For completeness, the commonly used \emph{Top-$K$ normalization} is
\begin{align}
    \tilde{g}_i(x) =
    \begin{cases}
        \dfrac{g_i(x)}{\sum_{j \in \mathcal{T}(x)} g_j(x)} & \text{if } i \in \mathcal{T}(x), \\
        0 & \text{otherwise}.
    \end{cases}
\end{align}
We also adopt the standard auxiliary load-balancing loss (used throughout our experiments) to encourage uniform expert utilization. Given a minibatch $\mathcal{B}$, define
\begin{align}
    f_i &= \frac{1}{|\mathcal{B}|}\sum_{x \in \mathcal{B}} \mathbb{I}[i \in \mathcal{T}(x)], \\
    p_i &= \frac{1}{|\mathcal{B}|}\sum_{x \in \mathcal{B}} s_i(x),
\end{align}
and compute
\begin{align}
    \mathcal{L}_\text{balance} = E \sum_{i=1}^{E} f_i \, p_i, \qquad
    \mathcal{L}_\text{total} = \mathcal{L}_\text{CE} + \lambda \mathcal{L}_\text{balance}.
\end{align}

\section{Extended Analysis on Related Work}
\label{append:related_work}

We notice a concurrent work~\citep{abnar2025parameters} studied scaling law for optimal MoE sparsity.
We highlight the differences between our work and theirs as follows:
\begin{itemize}
    \item \textbf{Formulation}: We define ``sparsity'' as the activation rate $r_\text{a}=\sfrac{N_\text{a}}{N}$, which is a more general definition than that proposed by \citet{abnar2025parameters}, namely the ratio of inactive experts to the total number of experts, $\sfrac{(E-K)}{E}$.
    \item \textbf{Methodology}: Given that the activation rate $r_\text{a}$ does not depend on the underlying model architecture, we can thus easily take into consideration other components such as shared expert and build all our models upon the optimized architecture proposed in \S~\ref{sec:arc}. This ensures the observed performance differences solely attribute to the varying activation rates.
    \item \textbf{Sufficient training}: Our main comparisons operate in a sufficiently trained regime (often beyond the compute-optimal point). 
    For example, the 2B MoE models in Table~\ref{tab:2B_keepD} have $\sfrac{D}{N}$ ranging from 53 to 252, 
    and the 7B MoE models in Table~\ref{tab:7B_keepD} have $\sfrac{D}{N}$ ranging from 20 to 93, 
    while our dense baselines satisfy $\sfrac{D}{N}\ge 20$ (Table~\ref{tab:dense_baseline}), 
    aligning with the common compute-optimal guideline~\citep{hoffmann2022training}. 
    This perspective is practically important since industrial model design often cares about the best achievable performance at a fixed-$N$ budget 
    (e.g., training a 7B model on trillions of tokens), 
    and it also enables more reliable downstream SFT comparisons that would be less meaningful with undertrained checkpoints.
    \item \textbf{Perspective under limited resources}: 
    Scaling-law studies often adopt broad sweeps that trade depth for breadth under limited compute, 
    which can lead to undertrained large-model settings (e.g., \citet{abnar2025parameters} uses $C\!=\!1\mathrm{e}20$ per run for a sweep up to 30B, 
    and \citet{ludziejewski2025jointmoescalinglaws} uses at most 80B tokens). 
    In contrast, our 7B study allocates substantially more compute per setting (e.g., multiple runs with $C\!\approx\!2.86\mathrm{e}21$ or $C\!\approx\!5.45\mathrm{e}21$; 
    Table~\ref{tab:7B_moe_ra} and Table~\ref{tab:dense_baseline}), 
    prioritizing a high-$\sfrac{D}{N}$ regime to more deeply investigate the fixed-$N$ \& fixed-$C$ question motivated by deployment memory constraints.
    \item \textbf{Conclusion}: We discover an optimal activation rate that appears to be \textit{independent} of model sizes, whereas~\citet{abnar2025parameters} find that the optimal sparsity increases with model size.
\end{itemize}
Our conclusion regarding a consistent optimal activation rate contradicts the findings of \citet{abnar2025parameters}.
While we believe our findings are reliable, given that our experiments are conducted with strictly controlled variables using \textit{optimized backbones} and \textit{sufficient training data}, we acknowledge the possibility that the optimal $r_\text{a}$ might slightly shift for model sizes significantly beyond our studied range (\ie, $N >> 7\text{B}$).
Nevertheless, we contend that the optimal $r_\text{a}$ can be considered consistent within a certain range of model sizes, in contrast to the significant changes reported by \citet{abnar2025parameters}.

\section{Pretraining Data Recipe}
\label{append:data_recipe}
For reproducibility, we provide the mixture ratios of our pretraining corpus and a comparison with the LLaMA-1 recipe~\citep{touvron2023llama1}.
Our recipe is intentionally close to a LLaMA-1--style mixture, and the corresponding data sources have public counterparts.

\begin{table*}[t]
    \centering
    \caption{Pretraining data recipe compared with the LLaMA-1 recipe.}
    \label{tab:data_recipe}
    \scriptsize
    \setlength\tabcolsep{3.5pt}
    \begin{tabular}{lccccr}
        \toprule
        DataSet Class & Our Recipe & Our Data Set Detail & LLaMA-1 Recipe & LLaMA-1 Data Set Detail & Recipe Diff \\
        \midrule
        WebData-en & 79.53\% & CC (English) & 82\% & 67\% CC + 15\% C4 (English) & -2.47\% \\
        Code & 4.62\% & The Stack & 4.50\% & Github-Big Query & +0.12\% \\
        Wikipedia & 5.06\% & en: 1.69\%, cn: 0.13\%, others: 3.24\% & 4.50\% & multi-lingual & +0.56\% \\
        Book & 5.18\% & open source English books & 4.50\% & book3, Gutenberg & +0.68\% \\
        arXiv & 3.38\% & as class name & 1.06\% & as class name & +2.32\% \\
        StackExchange & 2.21\% & as class name & 2.00\% & as class name & +0.21\% \\
        \bottomrule
    \end{tabular}
\end{table*}

%\section{Notation}\label{appendix-notation}

%Our notation is comprehensively summarized in Table~\ref{tab:notation}.

%\input{table/notation}

\section{Comprehensive List of Benchmarks}
\label{append:bmk}

To assess whether the optimal ARs generalize to downstream tasks, we conduct SFT on our 7B pre-trained models (trained w/ and w/o strict data reuse) and evaluate both the pre-trained models and SFT-ed models on a total number of 29 benchmarks (Figure~\ref{fig:downstream}). The comprehensive list of benchmarks is provided here.

For pre-trained models, we evaluate on:
\begin{itemize}
    \item Knowledge: BBH~\citep{suzgun2022challenging}, PIQA~\citep{bisk2019piqa}, SCIQ~\citep{welbl2017crowdsourcing}, SIQA~\citep{sap2019socialiqa} 
    \item Reasoning: ARC~\citep{clark2018think}, BoolQ~\citep{clark2019boolq}, CLUE~\citep{xu2020clue}, DROP~\citep{dua2019drop}, HellaSwag~\citep{zellers2019hellaswag}, NaturalQA~\citep{47761}, RACE~\citep{lai-etal-2017-race}, WinoGrande~\citep{sakaguchi2021winogrande}, XTREME~\citep{hu2020xtreme}
\end{itemize}
For SFT-ed models, we evaluate on:
\begin{itemize}
    \item Comprehensive: AGIEVAL~\citep{zhong2023agieval}, BBH
    \item Knowledge: CMMLU~\citep{li2023cmmlu}, MMLU~\citep{hendrycks2020measuring}, MMLU-Redux~\citep{gema2024we}, MMLU-Pro~\citep{wang2024mmlu}
    \item Reasoning: DROP, LiveBench~\citep{white2024livebench}, MuSR~\citep{sprague2023musr}
    \item Math: GAOKAO-Math24~\citep{zhang2023evaluating}, GSM8K~\citep{cobbe2021training}
    \item Code: APPS~\citep{hendrycks2021measuring}, DS-1000~\citep{lai2023ds}, HumanEval~\citep{chen2021evaluating}, LeetCode~\citep{coignion2024performance}, LiveCodeBench~\citep{jain2024livecodebench} 
\end{itemize}

\section{More Experimental Results}\label{appedix-results}

\begin{figure*}[th]
    \centering
    \includegraphics[width=\linewidth]{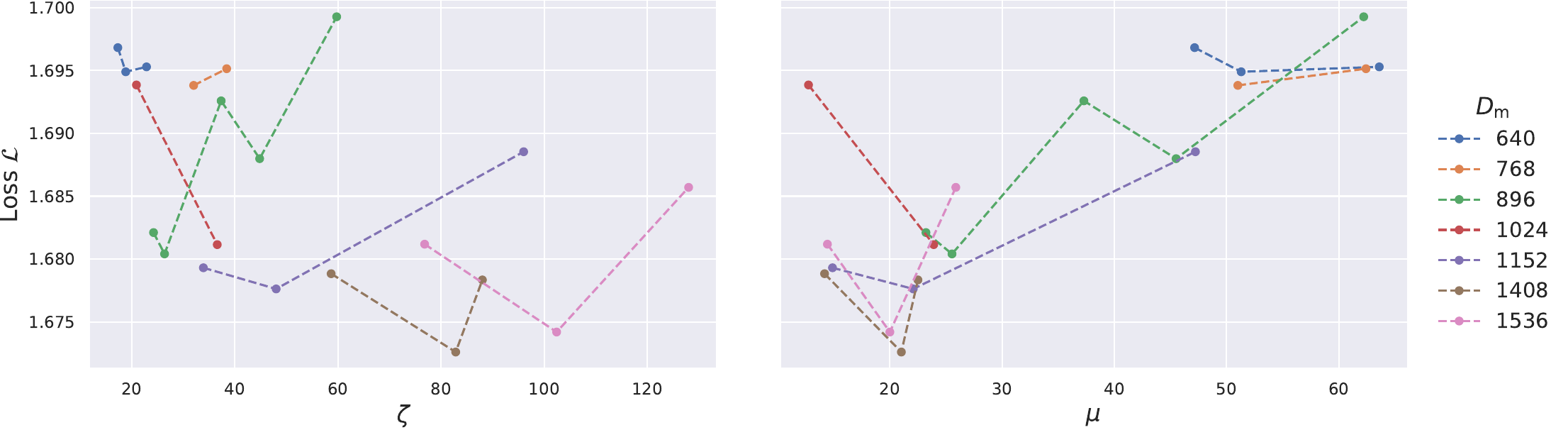}
    \caption{
        Results for model shape ratios $\zeta$ and $\mu$. An overall upward trend is observed in $\zeta$ as $D_\text{m}$ increases, while $\mu$ exhibits a downward trend with increasing $D_\text{m}$.
    }
    \label{fig:hidden_dimention}
\end{figure*}

\begin{table}[h]
    \centering
    \small
    \caption{Common training recipe.}
    \label{tab:training_recipe}
    \begin{tabular}{ll}
        \toprule
        Hyperparameter & Setting \\
        \midrule
        Vocab & 65536 \\
        Optimizer & Adam \\
        Weight decay & 0.1 \\
        Gradient clipping norm & 1.0 \\
        LR Scheduler & Cosine \\
        Warmup iters & $\operatorname{clip}(0.01 \cdot \text{Iters},200,2000) $ \\
        Min LR & 1e-5 \\
        \bottomrule
    \end{tabular}
\end{table}

% Requires: \usepackage{graphicx}
\begin{table*}[!t]
    \centering
    \scriptsize
    \setlength\tabcolsep{4pt}
    \caption{Experimental settings and results of MoE layer arrangement and shared expert. Hyperparameters shared by all experiments: $D_\text{m}=1408, D_{\text{ffn}}=3904, \operatorname{Norm}=\text{True}$.}
    \label{tab:layer_arrangement}
    \noindent\resizebox{\linewidth}{!}{\begin{tabular}{ccccccccccccc}
    \toprule
    $N$ & $N_\text{a}$ & $M$ &  $H$ & $D_\text{h}$ & $L$ & $E$ & $K$ & $D_\text{e}$ & $D_{\text{se}}$ & Scheme & $\mathcal{L}$ & Conclusion \\
    \midrule
    2.02B & 346M & 8.77e8 & 22 & 64 & 16 & 35 & 2 & 800 & 1600 & \texttt{full}+SE & 1.6813 & \multirow{3}{*}{\texttt{interleave} performs better than \texttt{full}} \\
    2.02B & 346M & 8.77e8 & 22 & 64 & 16 & 68 & 2 & 800 & 1600 & \texttt{interleave}+SE & 1.6766 \\
    2.02B & 346M & 8.77e8 & 22 & 64 & 16 & 70 & 4 & 800 & 0 & \texttt{interleave} & 1.6697 \\
    \midrule
    2.15B & 366M & 6.63e9 & 11 & 128 & 16 & 85 & 5 & 352 & 1760 & \texttt{1dense}+SE & 1.8700 & \multirow{4}{*}{\texttt{1dense}+SE performs the best} \\
    2.15B & 366M & 6.63e9 & 22 & 64 & 16 & 85 & 5 & 352 & 1760 & \texttt{1dense}+SE & 1.8557 \\
    2.15B & 367M & 6.63e9 & 11 & 128 & 16 & 70 & 4 & 800 & 0 & \texttt{interleave} & 1.8737 \\
    2.15B & 367M & 6.63e9 & 22 & 64 & 16 & 70 & 4 & 800 & 0 & \texttt{interleave} & 1.8620 \\
    \midrule
    2.15B & 368M & 9.31e8 & 22 & 64 & 17 & 37 & 4 & 800 & 0 & \texttt{1dense} & 1.6752 & \multirow{3}{*}{$\dfrac{D_\text{se}}{(D_\text{se}+KD_\text{e})}$ impacts little} \\
    2.15B & 368M & 9.31e8 & 22 & 64 & 17 & 36 & 3 & 800 & 800 & \texttt{1dense}+SE & 1.6712 \\
    2.15B & 368M & 9.31e8 & 22 & 64 & 17 & 35 & 2 & 800 & 1600 & \texttt{1dense}+SE & 1.6726 \\
    \bottomrule
    \end{tabular}}
\end{table*}

\begin{table*}[th]
    \centering
    \scriptsize
    \caption{Experimental settings and results of gate score normalization. Hyperparameters shared by all experiments: $\text{Scheme}=\texttt{1dense}, L=17, D_\text{m}=1408, D_{\text{ffn}}=3904, H=22, D_\text{h}=64$.}
    \label{tab:score_norm}
    \begin{tabular}{ccccccccccc}
        \toprule
        $N$ & $N_\text{a}$ & $r_\text{a}$ (\%) & $M$ & $E$ & $K$ & $D_\text{e}$ & $D_{\text{se}}$ & $\operatorname{Norm}$ & $\mathcal{L}$ & $\mathcal{\bar{L}_\text{balance}}$ \\
        \midrule
        2.15B & 368M & 17.08 & 9.31e8  & 35 & 2 & 800 & 1600 & Y & 1.6726 & \textbf{1.355} \\
        2.15B & 368M & 17.08 & 9.31e8 & 35 & 2 & 800 & 1600 & N & 1.6712 & 1.452 \\
        \midrule
        2.15B & 368M & 17.08 & 9.31e8 & 37 & 4 & 800 & 0 & Y & 1.6752 & \textbf{1.409} \\
        2.15B & 368M & 17.08 & 9.31e8 & 37 & 4 & 800 & 0 & N & 1.6750 & 1.440 \\
        \bottomrule
    \end{tabular}
\end{table*}

\begin{table*}[th]
    \centering
    \scriptsize
    \caption{Experimental settings and results of top-K setting. Hyperparameters shared by all experiments: $\text{Scheme}=\texttt{1dense}, L=16, D_\text{m}=1408, D_{\text{ffn}}=3904, H=11, D_\text{h}=128, \operatorname{Norm}=\text{False}$.}
    \label{tab:fine_grained}
    \begin{tabular}{ccccccccc}
        \toprule
        $N$ & $N_\text{a}$ & $r_\text{a}$  (\%) & $M$ & $E$ & $K$ & $D_\text{e}$ & $D_{\text{se}}$ & $\mathcal{L}$ \\
        \midrule
        % 2.15B & 368M & 17.08 & 9.31e8 & 17 & 35 & 2 & 800 & 1600 & 1.6771 \\
        % 2.15B & 366M & 17.04 & 9.17e8 & 16 & 85 & 5 & 352 & 1760 & \textbf{1.6700} \\
        % \midrule
        % 2.15B & 589M & 27.41 & 1.36e9 & 16 & 16 & 2 & 1760 & 3520 & 2.0353 \\
        % 2.15B & 589M & 27.41 & 1.36e9 & 16 & 88 & 11 & 320 & 3520 & \textbf{2.0330} \\
        2.15B & 591M & 27.47 & 8.00e9 & 8  & 1 & 3528 & 3528 & 2.0470 \\
        2.15B & 591M & 27.40 & 8.00e9 & 88 & 11 & 320 & 3520 & \textbf{2.0338} \\
        \midrule
        2.15B & 949M & 44.00 & 1.01e10 & 8  & 2 & 3176 & 6352 & \textbf{1.9996} \\
        2.15B & 948M & 44.05 & 1.01e10 & 88 & 22 & 288 & 6336 & 2.0266 \\
        \midrule
        2.15B & 1.24B & 57.57 & 1.19e10 & 8  & 3 & 2888 & 8664 & \textbf{2.0156} \\
        2.11B & 1.22B & 57.68 & 1.18e10 & 88 & 33 & 256 & 8448 & 2.0235 \\
        \bottomrule
    \end{tabular}
\end{table*}

\begin{table*}[!t]
    \centering
    \scriptsize
    \caption{Experimental settings and results of model shape ratios. Hyperparameters shared by all experiments: $\text{Scheme}=\texttt{1dense}, S=16384, D_\text{h}=128$.}
    \label{tab:hidden_dim}
    \begin{tabular}{cccccccccccccc}
    \toprule
    $N$ & $N_\text{a}$ & $D_\text{m}$ & $D_\text{ffn}$ & $L$ & $H$ & $E$ & $K$ & $D_\text{e}$ & $D_{\text{se}}$ & $\mu$ & $\zeta$ & $\mathcal{L}$ \\
    \midrule
    2.15e9 & 3.67e8 & 640 & 1774 & 34 & 5 & 50 & 4 & 608 & 2432 & 51.30 & 20.39 & 1.694 \\
    2.15e9 & 3.69e8 & 640 & 1774 & 37 & 5 & 38 & 3 & 736 & 2208 & 47.15 & 18.78 & 1.696 \\
    2.14e9 & 3.69e8 & 640 & 1774 & 49 & 5 & 41 & 3 & 512 & 1536 & 35.20 & 14.33 & 1.695 \\
    2.14e9 & 3.67e8 & 768 & 2129 & 20 & 6 & 99 & 8 & 448 & 3584 & 62.42 & 41.42 & 1.693 \\
    2.15e9 & 3.69e8 & 896 & 2484 & 15 & 7 & 124 & 10 & 416 & 4160 & 62.21 & 65.00 & 1.699 \\
    2.15e9 & 3.68e8 & 896 & 2484 & 20 & 7 & 91 & 7 & 416 & 2912 & 45.50 & 48.16 & 1.687 \\
    2.13e9 & 3.67e8 & 896 & 2484 & 24 & 7 & 54 & 4 & 576 & 2304 & 37.29 & 39.96 & 1.692 \\
    2.15e9 & 3.69e8 & 896 & 2484 & 34 & 7 & 61 & 4 & 352 & 1408 & 25.54 & 28.15 & 1.680 \\
    2.16e9 & 3.68e8 & 896 & 2484 & 37 & 7 & 47 & 3 & 416 & 1248 & 23.21 & 25.89 & 1.682 \\
    2.14e9 & 3.70e8 & 1024 & 2839 & 28 & 8 & 80 & 5 & 288 & 1440 & 23.91 & 38.93 & 1.681 \\
    2.16e9 & 3.69e8 & 1024 & 2839 & 49 & 8 & 49 & 2 & 256 & 512 & 12.75 & 22.33 & 1.693 \\
    2.15e9 & 3.67e8 & 1152 & 3194 & 12 & 9 & 79 & 6 & 640 & 3840 & 47.22 & 105.73 & 1.688 \\
    2.14e9 & 3.68e8 & 1152 & 3194 & 34 & 9 & 64 & 3 & 256 & 768 & 14.89 & 35.91 & 1.679 \\
    2.15e9 & 3.69e8 & 1280 & 3549 & 28 & 10 & 113 & 5 & 160 & 800 & 14.75 & 48.41 & 1.675 \\
    2.15e9 & 3.70e8 & 1408 & 3904 & 24 & 11 & 46 & 2 & 416 & 832 & 14.18 & 62.22 & 1.678 \\
    2.13e9 & 3.68e8 & 1536 & 4258 & 12 & 12 & 65 & 4 & 576 & 2304 & 25.88 & 140.64 & 1.685 \\
    2.16e9 & 3.68e8 & 1536 & 4258 & 15 & 12 & 91 & 5 & 320 & 1600 & 20.00 & 110.71 & 1.674 \\
    2.15e9 & 3.66e8 & 1536 & 4258 & 20 & 12 & 95 & 4 & 224 & 896 & 14.44 & 81.84 & 1.681 \\
    2.16e9 & 3.68e8 & 1792 & 4968 & 15 & 14 & 128 & 5 & 192 & 960 & 14.25 & 129.00 & 1.693 \\
    2.14e9 & 3.67e8 & 1920 & 5323 & 12 & 15 & 71 & 3 & 416 & 1248 & 16.03 & 175.55 & 1.699 \\
    \bottomrule
    \end{tabular}
\end{table*}

\begin{table*}[!t]
    \scriptsize
    \centering
    \caption{Experimental settings and results of optimal ARs for MoE models with $N=2.15\text{B}$ and fixed $r_\text{a}$. Hyperparameters shared by all experiments: $L=16, S=2048, D_\text{m}=1408, D_{\text{ffn}}=3904, H=11, D_\text{h}=128, \zeta=88$.}
    \label{tab:2B_keepD}
    \begin{tabular}{ccccccccccccccc}
    \toprule
    $N_\text{a}$ & $r_\text{a}$ (\%) & $M$ & $D$ & $C$ & $\sfrac{D}{N}$ & $E$ & $K$ & $D_\text{e}$ & $D_\text{se}$ & $\eta$ & $B$ & \# Iters & BPC \\
    \midrule
    1.88e8 & 8.74 & 1.68e9 & 1.14e11 & 1.92e20 & 53 & 89 & 1 & 352 & 352 & 2.01e-3 & 672 & 82833 & 0.5235 \\
    1.88e8 & 8.74 & 1.68e9 & 1.68e11 & 2.83e20 & 78 & 89 & 1 & 352 & 352 & 2.26e-3 & 832 & 98771 & 0.5159 \\
    1.88e8 & 8.74 & 1.68e9 & 3.67e11 & 6.16e20 & 170 & 89 & 1 & 352 & 352 & 2.87e-3 & 1344 & 133187 & 0.5090 \\
    1.88e8 & 8.74 & 1.68e9 & 5.41e11 & 9.10e20 & 252 & 89 & 1 & 352 & 352 & 3.24e-3 & 1728 & 152927 & 0.5048 \\
    \midrule
    2.33e8 & 10.81 & 1.95e9 & 1.14e11 & 2.22e20 & 53 & 88 & 2 & 352 & 704 & 2.01e-3 & 672 & 82833 & 0.5136 \\
    2.33e8 & 10.81 & 1.95e9 & 1.62e11 & 3.16e20 & 75 & 88 & 2 & 352 & 704 & 2.24e-3 & 896 & 88446 & 0.5084 \\
    2.33e8 & 10.81 & 1.95e9 & 2.31e11 & 4.50e20 & 107 & 88 & 2 & 352 & 704 & 2.49e-3 & 1024 & 110149 & 0.5027 \\
    2.33e8 & 10.81 & 1.95e9 & 3.29e11 & 6.41e20 & 153 & 88 & 2 & 352 & 704 & 2.78e-3 & 1280 & 125427 & 0.5002 \\
    2.33e8 & 10.81 & 1.95e9 & 4.68e11 & 9.12e20 & 218 & 88 & 2 & 352 & 704 & 3.10e-3 & 1600 & 142822 & 0.4967 \\
    \midrule
    4.11e8 & 19.11 & 3.02e9 & 1.14e11 & 3.44e20 & 53 & 84 & 6 & 352 & 2112 & 2.01e-3 & 672 & 82833 & 0.5013 \\
    4.11e8 & 19.11 & 3.02e9 & 1.46e11 & 4.42e20 & 68 & 84 & 6 & 352 & 2112 & 2.17e-3 & 768 & 93015 & 0.4971 \\
    4.11e8 & 19.11 & 3.02e9 & 1.88e11 & 5.67e20 & 87 & 84 & 6 & 352 & 2112 & 2.34e-3 & 960 & 95469 & 0.4953 \\
    4.11e8 & 19.11 & 3.02e9 & 2.41e11 & 7.27e20 & 112 & 84 & 6 & 352 & 2112 & 2.52e-3 & 1024 & 114870 & 0.4909 \\
    4.11e8 & 19.11 & 3.02e9 & 3.09e11 & 9.34e20 & 144 & 84 & 6 & 352 & 2112 & 2.73e-3 & 1280 & 117950 & 0.4872 \\
    \midrule
    7.52e8 & 34.95 & 5.06e9 & 1.14e11 & 5.77e20 & 53 & 84 & 15 & 320 & 4800 & 2.01e-3 & 672 & 82833 & 0.4963 \\
    7.52e8 & 34.95 & 5.06e9 & 1.80e11 & 9.13e20 & 84 & 84 & 15 & 320 & 4800 & 2.31e-3 & 896 & 98256 & 0.4892 \\
    \midrule
    1.09e9 & 50.79 & 7.11e9 & 1.14e11 & 8.10e20 & 53 & 84 & 26 & 288 & 7488 & 2.01e-3 & 672 & 82833 & 0.4950 \\
    1.09e9 & 50.79 & 7.11e9 & 1.28e11 & 9.13e20 & 60 & 84 & 26 & 288 & 7488 & 2.08e-3 & 704 & 89125 & 0.4933 \\
    \bottomrule
    \end{tabular}
\end{table*}

\begin{table*}[thbp]
    \scriptsize    
    \setlength\tabcolsep{5pt}
    \centering
    \caption{Experimental settings and results of optimal ARs for MoE models $N=2.15\text{B}$ with fixed $C$. Hyperparameters shared by all experiments: $L=16, S=2048, D_\text{m}=1408, D_{\text{ffn}}=3904, H=11, D_\text{h}=128, \zeta=88$. The green row corresponds to the MoE model with the lowest BPC on the validation set.}
    \label{tab:2B_moe_ra}
    \noindent\resizebox{\linewidth}{!}{\begin{tabular}{ccccccccccccccc}
    \toprule
    $N_\text{a}$ & $r_\text{a} (\%)$ & $M$ & $D$ & $C$ & $\sfrac{D}{N}$ & $\mu$ & $E$ & $K$ & $D_\text{e}$ & $D_\text{se}$ & $\eta$ & $B$ & \# Iters & BPC \\
    \midrule
    1.88e8 & 8.74  & 1.70e9 & 5.41e11 & 9.18e20 & 252 & 22.50 & 89 & 1 & 352 & 352 & 3.24e-3 & 1728 & 152927 & 0.5048 \\
    2.33e8 & 10.82 & 1.96e9 & 4.68e11 & 9.19e20 & 218 & 22.50 & 88 & 2 & 352 & 704 & 3.10e-3 & 1600 & 142822 & 0.4967 \\
    3.24e8 & 15.04 & 2.50e9 & 3.75e11 & 9.38e20 & 174 & 22.50 & 86 & 4 & 352 & 1408 & 2.89e-3 & 1344 & 136378 & 0.4896 \\
    3.68e8 & 17.11 & 2.77e9 & 3.39e11 & 9.38e20 & 158 & 22.50 & 85 & 5 & 352 & 1760 & 2.80e-3 & 1296 & 127756 & 0.4874 \\
    3.89e8 & 18.06 & 2.89e9 & 3.25e11 & 9.38e20 & 151 & 22.50 & 93 & 6 & 320 & 1920 & 2.76e-3 & 1280 & 123862 & 0.4871 \\
    4.11e8 & 19.12 & 3.03e9 & 3.09e11 & 9.38e20 & 144 & 22.50 & 84 & 6 & 352 & 2112 & 2.73e-3 & 1280 & 117950 & 0.4872 \\
    \rowcolor{green!15} 4.29e8 & 19.94 & 3.13e9 & 2.99e11 & 9.38e20 & 139 & 22.50 & 92 & 7 & 320 & 2240 & 2.70e-3 & 1248 & 117177 & \textbf{0.4857} \\
    5.90e8 & 27.46 & 4.10e9 & 2.23e11 & 9.15e20 & 104 & 22.55 & 8 & 1 & 3528 & 3528 & 2.46e-3 & 1024 & 106335 & 0.4907 \\
    7.52e8 & 34.96 & 5.08e9 & 1.80e11 & 9.16e20 & 84 & 22.50 & 84 & 15 & 320 & 4800 & 2.31e-3 & 896 & 98256 & 0.4892 \\
    9.48e8 & 44.11 & 6.25e9 & 1.47e11 & 9.16e20 & 68 & 22.56 & 8 & 2 & 3176 & 6352 & 2.16e-3 & 768 & 93460 & 0.4899 \\
    1.09e9 & 50.80 & 7.12e9 & 1.29e11 & 9.15e20 & 60 & 22.50 & 84 & 26 & 288 & 7488 & 2.08e-3 & 704 & 89125 & 0.4933 \\
    1.24e9 & 57.73 & 8.01e9 & 1.14e11 & 9.13e20 & 53 & 22.56 & 8 & 3 & 2888 & 8664 & 2.006e-3 & 672 & 82833 & 0.4934 \\
    \bottomrule
    \end{tabular}}
\end{table*}

\begin{table*}[!t]
    \scriptsize
    \centering
    \caption{Experimental settings and results of optimal ARs for MoE models with $N=6.52\text{B}$ with fixed $D$. Hyperparameters shared by all experiments: $L=24, S=2048, D_\text{m}=2048, D_{\text{ffn}}=5464, H=16, D_\text{h}=128, \zeta=85.3$.}
    \label{tab:7B_keepD}
    \noindent\resizebox{\linewidth}{!}{\begin{tabular}{cccccccccccccc}
    \toprule
    $N_\text{a}$ & $r_\text{a}$ (\%) & $M$ & $D$ & $C$ & $\sfrac{D}{N}$ & $E$ & $K$ & $D_\text{e}$ & $D_\text{se}$ & $\eta$ & $B$ & \# Iters & BPC \\
    \midrule
    7.26e8 & 11.15 & 5.59e9 & 1.30e11 & 7.25e20 & 19.90 & 82 & 2 & 512 & 1024 & 4.74e-4 & 640 & 98816 & 0.4808 \\
    8.70e8 & 13.36 & 6.46e9 & 1.30e11 & 8.37e20 & 19.90 & 81 & 3 & 512 & 1536 & 4.74e-4 & 640 & 98816 & 0.4763 \\
    1.02e9 & 15.67 & 7.33e9 & 1.30e11 & 9.49e20 & 19.90 & 80 & 4 & 512 & 2048 & 4.74e-4 & 640 & 98816 & 0.4737 \\
    1.31e9 & 20.03 & 9.07e9 & 1.30e11 & 1.17e21 & 19.88 & 78 & 6 & 512 & 3072 & 4.74e-4 & 640 & 98816 & 0.4703 \\
    1.70e9 & 26.11 & 1.15e10 & 1.30e11 & 1.48e21 & 19.90 & 86 & 10 & 448 & 4480 & 4.74e-4 & 640 & 98816 & 0.4681 \\
    3.47e9 & 53.30 & 2.21e10 & 1.30e11 & 2.86e21 & 19.90 & 84 & 28 & 384 & 10752 & 4.74e-4 & 640 & 98816 & 0.4664 \\
    \bottomrule
    \end{tabular}}
\end{table*}

\begin{table*}[thbp]
    \scriptsize
    \setlength\tabcolsep{5pt}
    \centering
    \caption{Experimental settings and results of optimal ARs for MoE models with $N=6.52\text{B}$ with fixed $C$. Hyperparameters shared by all experiments: $L=24, S=2048, D_\text{m}=2048, D_{\text{ffn}}=5464, H=16, D_\text{h}=128, \zeta=85.3$. The green row corresponds to the MoE model with the lowest BPC on the validation set.}
    \label{tab:7B_moe_ra}
    \noindent\resizebox{\linewidth}{!}{\begin{tabular}{ccccccccccccccc}
    \toprule
    $N_\text{a}$ & $r_\text{a} (\%)$ & $M$ & $D$ & $C$ & $\sfrac{D}{N}$ & $\mu$ & $E$ & $K$ & $D_\text{e}$ & $D_\text{se}$ & $\eta$ & $B$ & \# Iters & BPC \\
    \midrule
    5.85e8 & 8.97 & 4.73e9 & 6.05e11 & 2.86e21 & 92.88 & 21.00 & 83 & 1 & 512 & 512 & 7.62e-4 & 1512 & 195502 & 0.4665 \\
    7.30e8 & 11.19 & 5.59e9 & 5.11e11 & 2.86e21 & 78.47 & 21.00 & 82 & 2 & 512 & 1024 & 7.23e-4 & 1360 & 183630 & 0.4624 \\
    8.74e8 & 13.41 & 6.46e9 & 4.43e11 & 2.86e21 & 67.93 & 21.00 & 81 & 3 & 512 & 1536 & 6.92e-4 & 1232 & 175482 & 0.4580 \\
    1.02e9 & 15.63 & 7.33e9 & 3.90e11 & 2.86e21 & 59.89 & 21.00 & 80 & 4 & 512 & 2048 & 6.64e-4 & 1152 & 165447 & 0.4571 \\
    \rowcolor{green!15} 1.31e9 & 20.07 & 9.07e9 & 3.16e11 & 2.86e21 & 48.50 & 21.00 & 78 & 6 & 512 & 3072 & 6.23e-4 & 1040 & 148410 & \textbf{0.4543} \\
    1.71e9 & 26.18 & 1.15e10 & 2.50e11 & 2.86e21 & 38.32 & 21.00 & 86 & 10 & 448 & 4480 & 5.80e-4 & 960 & 127035 & 0.4580 \\
    1.96e9 & 30.07 & 1.30e10 & 2.21e11 & 2.86e21 & 33.83 & 21.00 & 84 & 12 & 448 & 5376 & 5.57e-4 & 800 & 134597 & 0.4588 \\
    3.48e9 & 53.38 & 2.21e10 & 1.30e11 & 2.86e21 & 19.87 & 21.00 & 84 & 28 & 384 & 10752 & 4.74e-4 & 640 & 98816 & 0.4670 \\
    \bottomrule
    \end{tabular}}
\end{table*}

\begin{table*}[!t]
    \scriptsize
    \centering
    \caption{Experimental settings and results of optimal ARs for 2B, 3B, and 7B dense baselines.}
    \label{tab:dense_baseline}
    \begin{tabular}{ccccccccccccc}
    \toprule
    $N$ & $M$ & $D$ & $C$ & $\sfrac{D}{N}$ & $L$ & $H$ & $D_\text{m}$ & $D_\text{ffn}$ & $\eta$ & $B$ & \# Iters & BPC \\
    \midrule
    2.15e9 & 1.44e10 & 6.50e10 & 9.36e20 & 30.23 & 28 & 17 & 2176 & 8848 & 8.44e-4 & 320 & 99182 & 0.4921 \\
    2.15e9 & 1.44e10 & 1.14e11 & 1.64e21 & 53.02 & 28 & 17 & 2176 & 8848 & 1.00e-3 & 448 & 124032 & 0.4808 \\
    \midrule 
    3.29e9 & 2.24e10 & 6.26e10 & 1.40e21 & 19.03 & 44 & 19 & 2432 & 7008 & 1.23e-3 & 448 & 68253 & 0.4833 \\
    3.29e9 & 2.24e10 & 1.25e11 & 2.80e21 & 38.06 & 44 & 19 & 2432 & 7008 & 1.52e-3 & 640 & 95554 & 0.4684 \\
    \midrule
    6.48e9 & 4.21e10 & 6.80e10 & 2.86e21 & 10.49 & 32 & 32 & 4096 & 11008 & 3.89e-4& 432 & 76813 & 0.4736 \\
    6.48e9 & 4.21e10 & 1.30e11 & 5.45e21 & 20.00 & 32 & 32 & 4096 & 11008 & 4.76e-4 & 640 & 98816 & 0.4594 \\
    \bottomrule
    \end{tabular}
\end{table*}

\begin{table*}[!t]
    \scriptsize
    \setlength\tabcolsep{4pt}
    \centering
    \caption{Experimental settings and results of data reusing ($\hat{D}=68\text{B}$) for MoE models with $N=6.52\text{B}$ with fixed $C$. Hyperparameters shared by all experiments: $L=24, S=2048, D_\text{m}=2048, D_{\text{ffn}}=5464, H=16, D_\text{h}=128, \zeta=85.3$. The green row corresponds to the MoE model with the lowest BPC on the validation set.}
    \label{tab:7B_data_reuse}
    \begin{tabular}{cccccccccccccccc}
    \toprule
    $N_\text{a}$ & $r_\text{a} (\%)$ & $M$ & $D$ & Epoch & $M$ & $\sfrac{D}{N}$ & $\mu$ & $E$ & $K$ & $D_\text{e}$ & $D_\text{se}$ & $\eta$ & $B$ & \# Iters & BPC \\
    \midrule
    7.30e8 & 11.19 & 5.59e9 & 5.11e11 & 7.52 & 2.86e21 & 78.47 & 21.00 & 82 & 2 & 512 & 1024 & 7.23e-4 & 1344 & 185816 & 0.4656 \\
    8.74e8 & 13.41 & 6.46e9 & 4.43e11 & 6.51 & 2.86e21 & 67.93 & 21.00 & 81 & 3 & 512 & 1536 & 6.92e-4 & 1232 & 175482 & 0.4618 \\
    1.02e9 & 15.63 & 7.33e9 & 3.90e11 & 5.74 & 2.86e21 & 59.89 & 21.00 & 80 & 4 & 512 & 2048 & 6.64e-4 & 1152 & 165447 & 0.4601 \\
    \rowcolor{green!15}  1.31e9 & 20.07 & 9.07e9 & 3.16e11 & 4.65 & 2.86e21 & 48.50 & 21.00 & 78 & 6 & 512 & 3072 & 6.23e-4 & 1024 & 150729 & \textbf{0.4590} \\
    1.71e9 & 26.18 & 1.15e10 & 2.50e11 & 3.67 & 2.86e21 & 38.32 & 21.00 & 86 & 10 & 448 & 4480 & 5.80e-4 & 960 & 127035 & 0.4597 \\
    1.96e9 & 30.07 & 1.30e10 & 2.21e11 & 3.24 & 2.86e21 & 33.83 & 21.00 & 84 & 12 & 448 & 5376 & 5.57e-4 & 792 & 135956 & 0.4603 \\
    \bottomrule
    \end{tabular}
\end{table*}

\begin{figure*}[!t]
    \centering
    \includegraphics[width=.62\linewidth]{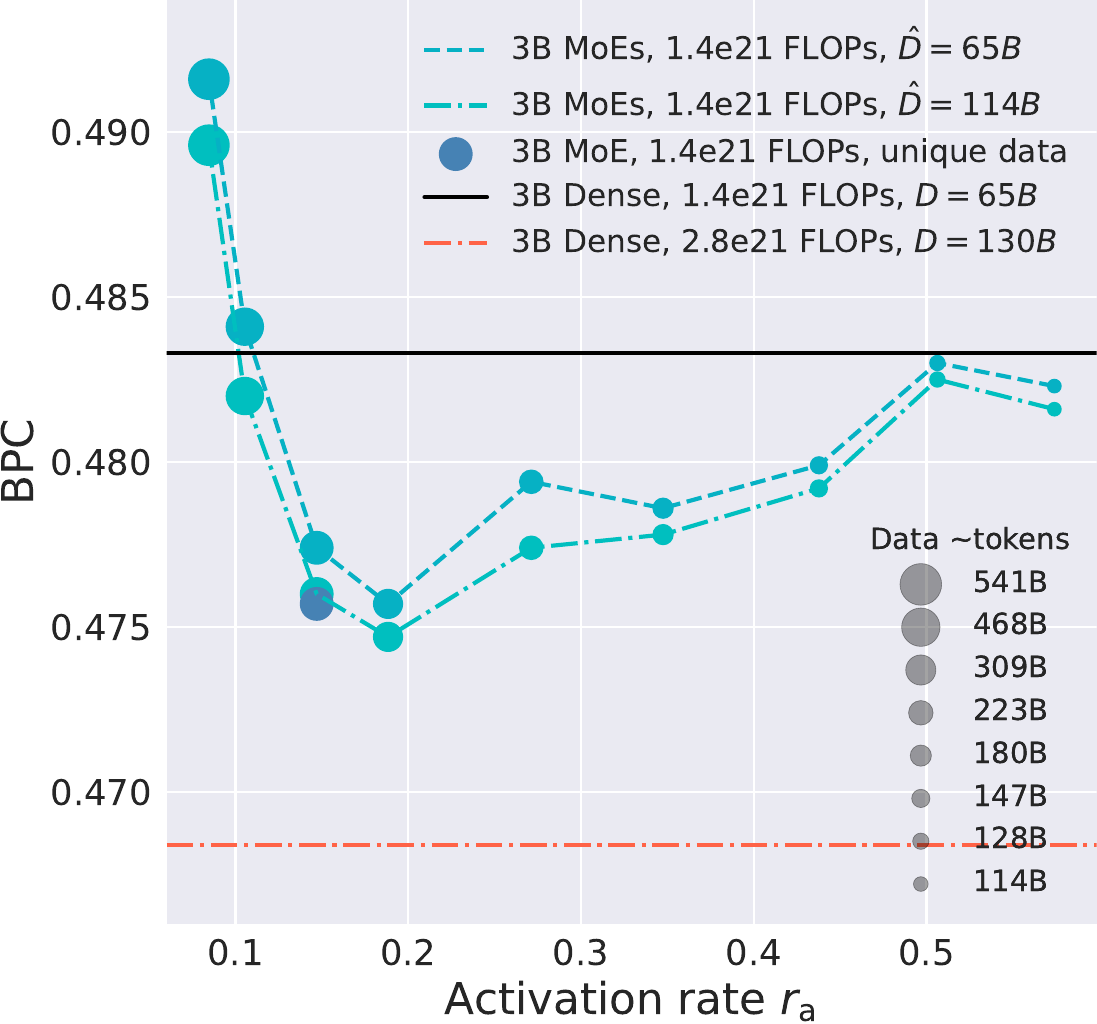}
    \caption{        
        Performance of $N \approx 3\text{B}$ models trained with varying data sizes $D$ and activation rate $r_\text{a}$.
        The optimal activation rate, $r_\text{a}^{**}=20\%$, aligns with the findings for the 2B models (Figure~\ref{fig:2B_optimal_ra}).
        Additionally, compared to training on the unique dataset, the data reuse scheme shows only a slight performance reduction.
        To save computational costs, only one model trained on unique data is included for reference.
        % We conclude three key findings for MoE models: 
        % 1) BPC decreases with increasing $D$.
        % 2) The performance gain does not depend linearly on the compute budget.
        % 3) An MoE model whose $r_\text{a}$ resides in the outstanding region $R_\text{a}^*$ is able to outperform its dense counterpart given the same compute $C$.
    }
    \label{fig:3B_optimal_ra}
\end{figure*}

\begin{table*}[!t]
    \scriptsize
    \setlength\tabcolsep{4.5pt}
    \centering
    \caption{
    Experimental settings and results of strict data reuse ($\hat{D}=65\text{B}$) for MoE models with $N=3.29\text{B}$ with fixed $C$. 
    Hyperparameters shared by all experiments: $L=24, S=2048, D_\text{m}=1408, D_{\text{ffn}}=3904, H=11, D_\text{h}=128$. The green row corresponds to the MoE model with the lowest BPC on the validation set.}
    \label{tab:3B_reuse_65B}
    \noindent\resizebox{\linewidth}{!}{\begin{tabular}{ccccccccccccccc}
    \toprule
    $N_\text{a}$ & $r_\text{a} (\%)$ & $D$ & Epoch & $M$ & $C$ & $\sfrac{D}{N}$ & $E$ & $K$ & $D_\text{e}$ & $D_\text{se}$ & $\eta$ & $B$ & \# Iters & BPC \\
    \midrule
    2.78e8 & 8.46 & 5.41e11 & 8.33 & 2.51e9 & 1.36e21 & 164.62 & 89 & 1 & 352 & 352 & 3.24e-3 & 1728 & 152927 & 0.4916 \\
    3.47e8 & 10.54 & 4.68e11 & 7.20 & 2.92e9 & 1.36e21 & 142.36 & 88 & 2 & 352 & 704 & 3.10e-3 & 1600 & 142822 & 0.4841 \\
    4.83e8 & 14.70 & 3.75e11 & 5.78 & 3.74e9 & 1.40e21 & 114.19 & 86 & 4 & 352 & 1408 & 2.89e-3 & 1344 & 136378 & 0.4774 \\
    \rowcolor{green!15} 6.20e8 & 18.83 & 3.09e11 & 4.76 & 4.56e9 & 1.41e21 & 94.06 & 84 & 6 & 352 & 2112 & 2.73e-3 & 1280 & 117950 & \textbf{0.4757} \\
    8.93e8 & 27.12 & 2.23e11 & 3.43 & 6.19e9 & 1.38e21 & 67.74 & 8 & 1 & 3528 & 3528 & 2.465e-3 & 1024 & 106335 & 0.4794 \\
    1.14e9 & 34.75 & 1.80e11 & 2.77 & 7.69e9 & 1.39e21 & 54.85 & 84 & 15 & 320 & 4800 & 2.31e-3 & 896 & 98256 & 0.4786 \\
    1.44e9 & 43.77 & 1.47e11 & 2.26 & 9.48e9 & 1.39e21 & 44.52 & 8 & 2 & 3176 & 6352 & 2.169e-3 & 768 & 93460 & 0.4799 \\
    1.66e9 & 50.63 & 1.29e11 & 1.98 & 1.08e10 & 1.39e21 & 39.09 & 84 & 26 & 288 & 7488 & 2.08e-3 & 704 & 89125 & 0.4830 \\
    1.89e9 & 57.40 & 1.14e11 & 1.75 & 1.22e10 & 1.39e21 & 34.61 & 8 & 3 & 2888 & 8664 & 2.006e-3 & 672 & 82833 & 0.4823 \\
    \bottomrule
    \end{tabular}}
\end{table*}

\begin{table*}[!t]
    \scriptsize
    \setlength\tabcolsep{4.5pt}
    \centering
    \caption{Experimental settings and results of data reuse ($\hat{D}=114\text{B}$) for MoE models with $N=3.29\text{B}$ with fixed $C$. Hyperparameters shared by all experiments: $L=24, S=2048, D_\text{m}=1408, D_{\text{ffn}}=3904, H=11, D_\text{h}=128$. The green row corresponds to the MoE model with the lowest BPC on the validation set.}
    \label{tab:3B_reuse_114B}
    \noindent\resizebox{\linewidth}{!}{\begin{tabular}{ccccccccccccccc}
    \toprule
    $N_\text{a}$ & $r_\text{a} (\%)$ & $D$ & Epoch & $M$ & $C$ & $\sfrac{D}{N}$ & $E$ & $K$ & $D_\text{e}$ & $D_\text{se}$ & $\eta$ & $B$ & \# Iters & BPC \\
    \midrule
    2.78e8 & 8.46 & 5.41e11 & 4.75 & 2.51e9 & 1.36e21 & 164.62 & 89 & 1 & 352 & 352 & 3.24e-3 & 1728 & 152927 & 0.4896 \\
    3.47e8 & 10.54 & 4.68e11 & 4.11 & 2.92e9 & 1.36e21 & 142.36 & 88 & 2 & 352 & 704 & 3.10e-3 & 1600 & 142822 & 0.4820 \\
    4.83e8 & 14.70 & 3.75e11 & 3.29 & 3.74e9 & 1.40e21 & 114.19 & 86 & 4 & 352 & 1408 & 2.89e-3 & 1344 & 136378 & 0.4760 \\
    \rowcolor{green!15} 6.20e8 & 18.83 & 3.09e11 & 2.71 & 4.56e9 & 1.41e21 & 93.93 & 84 & 6 & 352 & 2112 & 2.73e-3 & 1280 & 117950 & \textbf{0.4747} \\
    8.93e8 & 27.12 & 2.23e11 & 1.96 & 6.19e9 & 1.38e21 & 67.74 & 8 & 1 & 3528 & 3528 & 2.465e-3 & 1024 & 106335 & 0.4774 \\
    1.14e9 & 34.75 & 1.80e11 & 1.58 & 7.69e9 & 1.39e21 & 54.85 & 84 & 15 & 320 & 4800 & 2.31e-3 & 896 & 98256 & 0.4778 \\
    1.44e9 & 43.77 & 1.47e11 & 1.29 & 9.48e9 & 1.39e21 & 44.52 & 8 & 2 & 3176 & 6352 & 2.169e-3 & 768 & 93460 & 0.4792 \\
    1.66e9 & 50.63 & 1.29e11 & 1.13 & 1.08e10 & 1.39e21 & 39.09 & 84 & 26 & 288 & 7488 & 2.08e-3 & 720 & 87144 & 0.4825 \\
    1.89e9 & 57.40 & 1.14e11 & 1.00 & 1.22e10 & 1.39e21 & 34.61 & 8 & 3 & 2888 & 8664 & 2.006e-3 & 672 & 82833 & 0.4816 \\
    \bottomrule
    \end{tabular}}
\end{table*}

\begin{table}[h]
    \scriptsize
    \setlength\tabcolsep{5.5pt}
    \centering
    \caption{Experimental settings and results of loose data reuse for MoE models with $N=6.52\text{B}$ with fixed $C$. Hyperparameters shared by all experiments: $L=24, S=2048, D_\text{m}=2048, D_{\text{ffn}}=5464, H=16, D_\text{h}=128$. The green row corresponds to the MoE model with the lowest BPC on the validation set.}
    \label{tab:7B_loose_reuse}
    \begin{tabular}{cccccccccccccc}
    \toprule
    $N_\text{a}$ & $r_\text{a} (\%)$ & $\hat{D}$ & $M$ & $C$ & $\sfrac{D}{N}$ & $E$ & $K$ & $D_\text{e}$ & $D_\text{se}$ & $\eta$ & $B$ & \# Iters & BPC \\
    \midrule
    7.30e8 & 11.19 & 2.56e11 & 5.59e9 & 2.86e21 & 78.47 & 82 & 2 & 512 & 1024 & 7.23e-4 & 1344 & 185816 & 0.4591 \\
    8.74e8 & 13.41 & 2.21e11 & 6.46e9 & 2.86e21 & 67.93 & 81 & 3 & 512 & 1536 & 6.92e-4 & 1232 & 175482 & 0.4557 \\
    1.02e9 & 15.63 & 1.95e11 & 7.33e9 & 2.86e21 & 59.89 & 80 & 4 & 512 & 2048 & 6.64e-4 & 1152 & 165447 & 0.4550 \\
    \rowcolor{green!15} 1.31e9 & 20.07 & 1.58e11 & 9.07e9 & 2.87e21 & 48.50 & 78 & 6 & 512 & 3072 & 6.23e-4 & 1024 & 150729 & \textbf{0.4549} \\
    1.71e9 & 26.18 & 1.25e11 & 1.15e10 & 2.86e21 & 38.32 & 86 & 10 & 448 & 4480 & 5.80e-4 & 960 & 127035 & 0.4570 \\
    1.96e9 & 30.07 & 1.10e11 & 1.30e10 & 2.86e21 & 33.83 & 84 & 12 & 448 & 5376 & 5.57e-4 & 792 & 135956 & 0.4583 \\
    \bottomrule
    \end{tabular}
\end{table}

\clearpage
\section{The Use of Large Language Models: An Explanation}
Only a small fraction of complex paragraphs are written with the assistance and modification of ChatGPT. For instance, we provide the prompt: ``I am writing an academic conference paper in the field of computer science. Please help me polish the wording of this paragraph, organize the sentences, and express them in a more academic way.'' After that the outputs are rigorously reviewed to ensure accuracy and appropriateness.

\end{document}